\documentclass[10pt,journal,compsoc]{IEEEtran}

\usepackage{amsmath,amsfonts,amssymb}
\usepackage{array}
\usepackage[caption=false,font=normalsize,labelfont=sf,textfont=sf]{subfig}
\usepackage{textcomp}
\usepackage{stfloats}
\usepackage{url}
\usepackage{graphicx}
\usepackage{cite}
\usepackage{algorithm}
\usepackage{algorithmic}

\begin{document}

\title{Object Concepts Emerge from Motion}

\author{Boshi~Li, Xiaohui~Wang, Xiaoyang~Wu, Zhichao~Li, Ya~Yang, and Naiyan~Wang%
\IEEEcompsocitemizethanks{%
\IEEEcompsocthanksitem B. Li and Y. Yang are with Beijing University of Posts
and Telecommunications (e-mail: liboshi@bupt.edu.cn; yangya@bupt.edu.cn).
\IEEEcompsocthanksitem X. Wu is with the Department of Computer Science,
The University of Hong Kong, Hong Kong (e-mail: xiaoyang.wu@connect.hku.hk).
\IEEEcompsocthanksitem X. Wang, Z. Li, and N. Wang are with Xiaomi EV,
Beijing, China (e-mail: wangxiaohui.cs@gmail.com; leeisabug@gmail.com;
winsty@gmail.com).
N. Wang is the corresponding author.}%
\thanks{
\protect\par
% \protect\vspace{0.5\baselineskip}%
Project page: \protect\url{https://tj12342.github.io/object-concepts-from-motion/}}
}

\markboth{IEEE Transactions on Pattern Analysis and Machine Intelligence}%
{Li \MakeLowercase{\textit{et al.}}: Object Concepts Emerge from Motion}

\IEEEtitleabstractindextext{%

\begin{center}
\setlength{\tabcolsep}{1pt}
\begin{tabular}{@{}ccccc@{}}
  \includegraphics[width=0.16\textwidth]{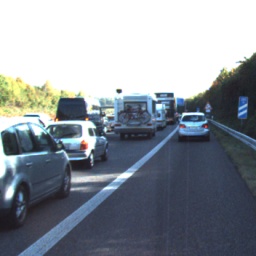} &
  \includegraphics[width=0.16\textwidth]{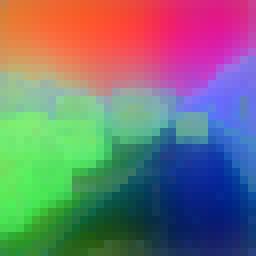} &
  \includegraphics[width=0.16\textwidth]{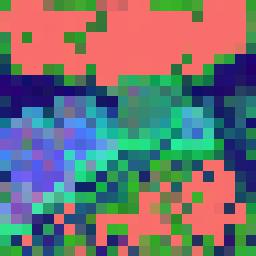} &
  \includegraphics[width=0.16\textwidth]{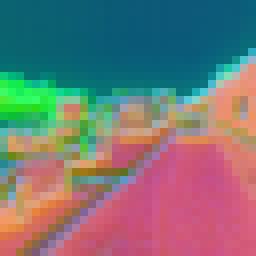} &
  \includegraphics[width=0.16\textwidth]{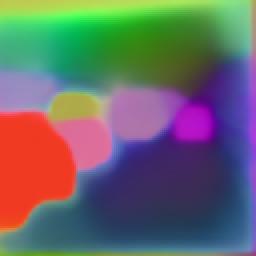} \\
  \includegraphics[width=0.16\textwidth]{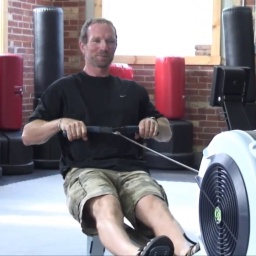} &
  \includegraphics[width=0.16\textwidth]{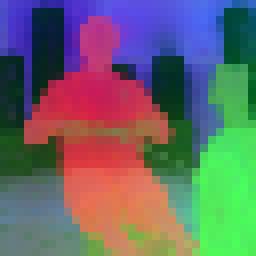} &
  \includegraphics[width=0.16\textwidth]{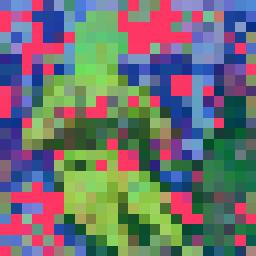} &
  \includegraphics[width=0.16\textwidth]{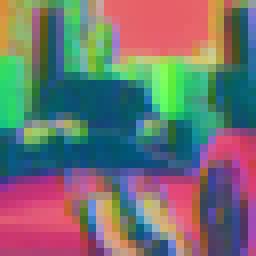} &
  \includegraphics[width=0.16\textwidth]{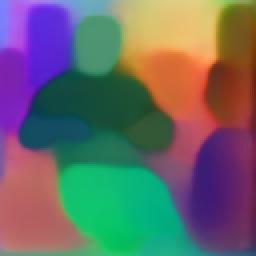} \\
  Original Image & DINOv3 & CLIP & MAE & Ours
\end{tabular}
\par\vspace{0.4em}
{\small\textbf{Fig.~1.} Motion teaches a single-image encoder which pixels belong to each object, internalizing dynamic grouping into static features. Compared with DINOv3, CLIP, and MAE, ours preserves within-object coherence while separating distinct instances. This instance-aware structure supports geometry and instance-sensitive scene understanding.}
\end{center}
\vspace{0.5em}
\setcounter{figure}{1}

\begin{abstract}
Object concepts play a foundational role in human visual cognition, enabling perception, memory, and interaction in the physical world. Motion, in turn, provides complementary cues that are not directly available from static appearance alone. Inspired by findings in developmental neuroscience, where infants acquire object understanding through the observation of motion, we propose a biologically inspired framework for learning object-centric visual representations from raw videos. Our key insight is that motion boundaries provide a strong signal for object-level grouping and can be used to derive pseudo-instance supervision. Concretely, we generate motion-derived instance masks using off-the-shelf optical flow methods and clustering, and use them to train a single-image encoder through pixel-level pairwise metric learning. Our framework requires no human annotations or camera calibration, making it scalable to large-scale unstructured video data. 
We first extract 195M pseudo-labeled frames from 7,163 hours of video and then expand this set to 421M frames through a second-stage Motion-Verified Self-Training paradigm. We use these frames to train a visual encoder up to Swin-H and further distill it into a family of Swin backbones.
%We scale motion-derived pretraining to 195M pseudo-labeled frames and introduce Motion-Verified Self-Training, which combines model proposals with motion cues to expand supervision to 421M pseudo-labeled frames from the same videos. We further scale the encoder to Swin-H and distill it into a family of Swin backbones.
We evaluate the learned representations on four downstream tasks spanning low- and high-level vision: monocular depth estimation, 3D object detection, 3D occupancy prediction, and end-to-end planning. Our models achieve competitive or superior performance compared to supervised and self-supervised pretraining baselines, with particularly strong results on geometry- and instance-sensitive tasks. These results suggest that motion-induced object representations provide a distinct perspective for visual pretraining, capturing a crucial but overlooked abstraction: the visual instance.
\end{abstract}

\begin{IEEEkeywords}
Object-centric learning, self-supervised learning, optical flow, video pretraining, visual foundation models.
\end{IEEEkeywords}
}

\maketitle
\IEEEdisplaynontitleabstractindextext
\IEEEpeerreviewmaketitle

\section{Introduction}
\label{sec:introduction}
\IEEEPARstart{P}{hysical} AI aims to develop intelligent agents capable of perceiving and interacting with the physical world. A fundamental cognitive capacity required for such agents is the ability to recognize and understand the concept of "object"—a core unit of perception and reasoning. In the human visual system, the importance of object concepts is well-established in neuroscience. As noted by Kellman and Spelke~\cite{kellman1983perception}, {\it ``this cognitive ability not only supports object recognition and classification, but also plays a crucial role in spatial perception, memory formation, and the interaction between objects and their environment.''} Understanding how object concepts are formed and represented in biological systems provides critical insights for building more robust and generalizable visual agents in artificial systems.

However, {\it what makes an object look like an object?} This is a non-trivial question, as objects can vary drastically in appearance, shape, and motion patterns.
Early studies in developmental neuroscience~\cite{kellman1983perception} have demonstrated that the ability to perceive object unity is not innate, but learned during infancy. Infants begin to exhibit evidence of understanding object cohesion from around two months of age, with robust performance observed by four months. These findings suggest that object perception is a learned capacity grounded in sensory experience.
Subsequent research~\cite{johnson2003development,carter_smith2003motion} has shown that motion cues—particularly common or coherent motion—serve as a powerful signal for infants to infer object boundaries and unity. As the visual system matures, this dynamic understanding is gradually internalized into the ventral visual stream~\cite{kravitz2011visuospatial,grill2001lateral}, enabling object recognition from static visual inputs alone.
Inspired by this developmental trajectory, our work aims to design an unsupervised computational model that mimics this learning process: {\it beginning from motion-based interactions and evolving toward abstract, appearance-based object concepts.}

Recently, learning universal visual representations through self-supervised or weakly supervised paradigms has gained significant attention due to their strong performance across a wide range of vision tasks. Among self-supervised approaches, notable examples include the DINO~\cite{caron2021emerging, oquab2023dinov2} and MAE~\cite{he2022masked, xie2022simmim} families, which rely on self-distillation and self-reconstruction mechanisms, respectively, to learn robust feature representations. Another influential direction leverages web-scale image-text pairs, as exemplified by CLIP~\cite{radford2021learning}, to align visual and language representations.
To better understand what these models capture, we compare the low-dimensional PCA projections of features extracted by DINO, CLIP, and our model (see Fig.~1). We observe that DINO and CLIP tend to focus on semantic {\it categories}. However, neither method captures the concept of a semantic {\it instance}—a distinct, coherent object entity—adequately. We argue that existing visual foundation models overlook this crucial level of abstraction, which is fundamental for understanding the physical world.

In this work, we propose a biologically inspired framework for learning visual features that encode object-level semantics. The key observation that inspires our method is that motion boundaries often align with object boundaries (detailed in Sec.~\ref{sec:motion_supervision}), which echoes the discoveries in neuroscience that common motion is crucial to the early development of object unity.
Based on this observation, we employ an off-the-shelf optical flow estimation algorithm, followed by a simple clustering technique, to generate pseudo-instance masks without human supervision. These instance labels are then used to supervise representation learning via a contrastive objective. Importantly, unlike previous approaches~\cite{zhou2017unsupervised,bian2019unsupervised}, our method does not require camera calibration parameters, allowing it to scale to large and diverse unlabeled video datasets.

A preliminary version of this work~\cite{liang2025object} introduced the motion-derived pretraining framework on 48M pseudo-labeled frames from driving videos. This article extends the framework around a single question: \emph{how does motion-derived object supervision scale?} We study this question along three axes: the size and diversity of the video corpus, the amount of reliable supervision extracted from each video, and the capacity of the pretrained model.

 \emph{Scaling the data.} A unified processing pipeline produces 195M (vs.\ 48M in the conference version) frames with motion-derived pseudo-labels from 7,163 hours of heterogeneous driving and self-collected web videos. \emph{Scaling the supervision.} Inspired by the data-engine principle of SAM~2~\cite{ravi2024sam2}, Motion-Verified Self-Training combines proposals from the initial pretrained encoder with motion evidence discarded during stage-1 label generation, further expanding usable supervision to 421M pseudo-labeled frames from the same video sources. \emph{Scaling the model.} We train the model up to Swin-H (vs.\ Swin-L in the conference version) and then distill its representation into a family of Swin backbones.

The extended framework achieves strong performance compared to supervised and self-supervised baselines on four downstream low- and high-level tasks, spanning from depth estimation to end-to-end driving.
% On KITTI depth estimation, our Swin-L obtains 0.042 Abs Rel, 1.715 RMSE, and 0.988 $\delta_1$, while the Swin-H encoder reaches 0.043 Abs Rel and 1.712 RMSE; on nuScenes 3D detection, Swin-L reaches 55.89 NDS and 47.59 mAP, while Swin-H reaches 56.92 NDS and 48.87 mAP; our models also transfer competitively to 3D occupancy and end-to-end planning under NAVSIMv2. The expanded pseudo-label coverage and the downstream results demonstrate the effectiveness of the complete Cycle-2 pipeline. A controlled Swin-T comparison further strengthens this conclusion: with a slightly smaller additional sample-exposure budget, 10 epochs on the Cycle-2 labels outperform extending Cycle-1 pretraining from 50 to 75 epochs across all non-saturated depth metrics. This controlled result demonstrates that switching to the improved Cycle-2 supervision yields better transfer than simply training longer on the original labels.
Our contributions are summarized as follows:
\begin{itemize}
\item We present a biologically inspired paradigm for object-centric visual representation learning that converts optical flow into pseudo-instance supervision and requires neither human annotations on the target video corpus nor camera calibration, and we demonstrate its effectiveness and scalability on modern architectures.
\item We scale motion-derived supervision to 195M pseudo-labeled frames from 7,163 hours of heterogeneous driving and self-collected videos through a unified large-scale processing pipeline.
\item We propose Motion-Verified Self-Training, which combines model proposals from initial training with independent motion evidence to expand usable supervision to 421M pseudo-labeled frames from the same video sources. We evaluate the resulting complete Cycle-2 pipeline, while a controlled Swin-T experiment isolates the benefit of the Cycle-2 supervision relative to continued pretraining on the original labels.
\item We scale the backbone to Swin-H, distill it into a family of Swin backbones, and extensively evaluate the resulting models on four downstream tasks: monocular depth estimation, 3D object detection, 3D occupancy prediction, and end-to-end planning. Our models achieve competitive or superior performance relative to other baselines.
\end{itemize}

%Relative to that conference version, this journal article scales motion-derived pretraining from 48M to 195M frames, introduces Motion-Verified Self-Training to obtain 421M pseudo-labeled frames (Sec.~\ref{sec:motion_verified_self_training}), scales the backbone to Swin-H with compact-model distillation, and adds end-to-end planning and broader experimental analyses (Sec.~\ref{sec:experiments}).

\section{Related Work}
\label{sec:related_work}

\subsection{Object Discovery}
\label{sec:rw_object_discovery}
Object discovery aims to identify and localize objects without instance-level annotations for predefined categories. Early methods exploit repeated object occurrence across images~\cite{joulin2010discriminative,joulin2012multi,vicente2011object} or select object regions from proposal sets through combinatorial optimization~\cite{uijlings2013selective,vo2019unsupervised,vo2021large,wei2019unsupervised,zitnick2014edge}. More recent approaches use features from pretrained image encoders, such as DINO, together with graph partitioning or spectral clustering to localize and segment objects~\cite{simeoni2021localizing,wang2022self,wang2023tokencut,zhang2024heap}. Object-centric generative models instead decompose a scene into entities by learning to reconstruct or generate its constituent objects and background~\cite{burgess2019monet,greff2019multi,locatello2020object,luo2024unsupervised}. Category-agnostic discovery has also been extended to jointly produce object detections and instance masks without manual category labels~\cite{wang2023cutlearn}.

Motion and multimodal observations provide another route to object discovery. These methods use the motion consistency of 2D or 3D points to separate objects from their surroundings~\cite{singh2023locate,wang20224d,yang2021dystab,xie2022moseg}. Our work also extracts object regions from motion, but object discovery is not the final task. We use optical flow and clustering to construct category-agnostic pseudo-instance masks at scale, which then supervise a single-image representation. This setup separates video motion, the source of supervision, from the deployed encoder, which requires only a static image.

\subsection{Learning Representations from Motion}
\label{sec:rw_motion_representation}
Motion can supervise visual representation learning without requiring motion at inference time. Early work used ego-motion as a supervisory signal, connecting visual features to the movement of the observer~\cite{agrawal2015learning}. Pathak~et~al.~\cite{pathak2017learning} obtained foreground masks from unsupervised motion segmentation and trained a ConvNet to predict them, demonstrating that motion-derived grouping can transfer to recognition. DyStaB~\cite{yang2021dystab} likewise bootstraps a static object model from dynamic segmentation, while MoSeg~\cite{xie2022moseg} and Divided Attention (DivA)~\cite{lao2025divided} discover multiple moving regions with object-centric motion models and use them to support representation learning. We share this motion-to-representation principle, but use motion-derived multi-instance masks as dense pairwise supervision, directly organizing pixel features around object unity and instance separation. %Our pipeline requires neither ego-motion measurements nor camera calibration, scales to 421M pseudo-labeled frames and modern hierarchical backbones, and uses motion-verified self-training to recover supervision discarded by conservative flow clustering.

Motion is also central to unsupervised video object segmentation, where the objective is to produce object masks from video. Representative methods discover moving objects through adversarial contextual reasoning~\cite{yang2019unsupervised}, appearance-motion decomposition~\cite{liu2021emergence}, slot-based flow grouping~\cite{yang2021self}, or motion anticipation~\cite{choudhury2022guess}. In our framework, the masks are intermediate supervision rather than the final output: their effectiveness is evaluated by transferring the features to downstream applications, not by evaluating the predicted masks directly.

Video self-supervised methods such as VideoMAE~\cite{tong2022videomae} and V-JEPA~\cite{bardes2024revisiting} learn spatio-temporal representations by reconstructing or predicting masked video content. They capture motion implicitly through temporal context and typically train encoders designed for video clips. In contrast, we explicitly convert motion into object-level grouping supervision and train an encoder that operates on a single image at inference time.

\subsection{Visual Foundation Models}
\label{sec:rw_foundation_models}
Visual foundation models learn transferable representations from large-scale data using several forms of pretraining. Contrastive methods enforce invariance across augmented views~\cite{chen2020simple,chen2020improved,chen2021empirical,dwibedi2021little,he2020momentum,yeh2022decoupled}, including dense pixel- and object-level variants that explicitly preserve local correspondence or object structure~\cite{wang2021densecl,xie2021pixpro,henaff2021detcon}, while self-distillation methods obtain supervisory targets from a teacher network without requiring negative pairs~\cite{caron2021emerging,darcet2023vision,grill2020bootstrap,oquab2023dinov2,simeoni2025dinov3}. Masked-image modeling learns by reconstructing missing image content~\cite{bao2021beit,he2022masked,peng2022beit,wang2023image,xie2022simmim}, and joint embedding-prediction objectives predict representations rather than pixels~\cite{assran2023self,bardes2024revisiting}. Vision-language models such as CLIP and the SigLIP series additionally exploit image-text supervision to acquire broad category semantics~\cite{radford2021learning,zhai2023sigmoid,tschannen2025siglip2}.

Large-scale pretraining has also been specialized for structured visual outputs. Promptable segmentation models such as SAM and SAM~2 learn broadly applicable mask prediction from large segmentation corpora~\cite{kirillov2023segment,ravi2024sam2}. Our method instead uses motion as a category-agnostic grouping signal, encouraging pixels within an object to remain coherent and adjacent instances to remain distinguishable. As suggested by Fig.~1, this inductive bias differs from objectives dominated by category invariance or appearance reconstruction, and is particularly relevant to geometry- and instance-sensitive downstream tasks.

\subsection{Self-Training and Pseudo-Label Scaling}
\label{sec:rw_self_training}
Self-training exploits unlabeled data by using a seed model to generate pseudo-labels for subsequent training~\cite{lee2013pseudo,xie2020self}. Confidence filtering, consistency regularization, and teacher-student targets can improve the reliability of this process~\cite{tarvainen2017mean,sohn2020fixmatch}. Quality control is especially important for dense supervision: boundary errors, merged instances, and missing object parts affect many pixels and can be amplified when model predictions are reused as training targets. Dynamic-static bootstrapping further shows that motion-derived uncertainty can constrain feedback from a learned static model~\cite{yang2021dystab}.

Large segmentation systems address this issue through iterative data engines. SAM uses a model-assisted annotation loop to construct a large mask corpus~\cite{kirillov2023segment}. SAM~2 extends this strategy to video: progressively stronger models assist human annotators, propose additional masks, and help determine which annotations require correction before entering the training set~\cite{ravi2024sam2}. The broader lesson is that scaling pseudo-labels requires a verification signal rather than accepting model predictions wholesale.

Our Motion-Verified Self-Training adopts this principle without manual verification on the target video corpus. Model proposals increase mask coverage, while proposal-independent motion evidence verifies and refines them. Motion therefore remains an external constraint on the self-training loop, distinguishing our approach from conventional self-training based only on prediction confidence or teacher-student agreement.

\section{Method}
\label{sec:method}
In this section, we first briefly introduce our training pipeline and then present a theoretical analysis of why flow boundaries indicate object boundaries. Next, we elaborate on the data processing of our two training stages and the training details.

\subsection{Overview}
\label{sec:method_overview}
Our method learns an object-centric image representation from raw videos without manual instance annotations. The central supervision signal is motion: coherent motion suggests object unity, while motion discontinuities often reveal object boundaries. Motion is used only during pretraining; after training, the encoder takes a single image as input and serves as a standard visual backbone.

\begin{figure*}[t]
\centering
\includegraphics[width=0.98\textwidth]{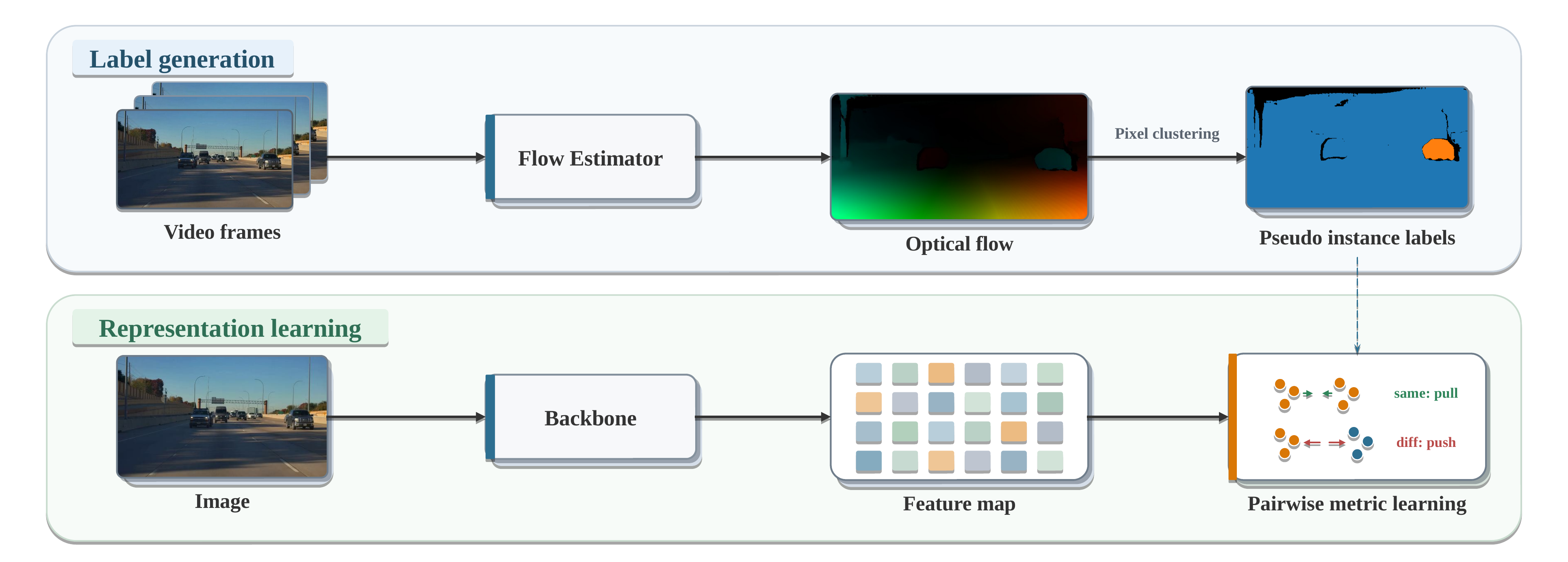}
\caption{Cycle-1 motion-derived object supervision. The first row converts raw video frames into pseudo-instance labels: a flow estimator produces optical flow, and pixel clustering over reliable flow regions yields pseudo-instance labels. The second row trains an image encoder from these labels: a single image is encoded into dense features and optimized with pairwise metric learning, pulling pixels from the same pseudo-instance together and pushing pixels from different instances apart.}
\label{fig:motion_supervision_pipeline}
\end{figure*}

We design a two-cycle training paradigm. In Cycle-1, shown in Fig.~\ref{fig:motion_supervision_pipeline}, we convert optical flow into motion-derived pseudo-instance labels and train dense image features with a pairwise metric-learning objective. We scale this supervision across heterogeneous video sources with a unified processing and filtering pipeline, yielding 195M pseudo-labeled frames. In this stage, we generate labels only over highly reliable regions to ensure the correctness of the initial supervision. In Cycle-2, shown in Fig.~\ref{fig:motion_verified_self_training}, we propose Motion-Verified Self-Training to better utilize the low-confidence motion cues discarded in Cycle-1: the Cycle-1 encoder proposes candidate masks, while motion cues verify and refine them into Cycle-2 pseudo-labels. We train a Cycle-2 Swin-H teacher on the resulting 421M motion-verified pseudo-labeled frames and subsequently distill its features into all deployed backbones, including Swin-H. The complete Cycle-2 pipeline is evaluated in Sec.~\ref{sec:exp_mvst_effect}.

\subsection{Motion-Derived Object Supervision}
\label{sec:motion_supervision}
Motion boundaries provide a geometric cue for object boundaries. When an object moves independently, its image motion is separate from the surrounding environment. Ego-motion can also turn depth discontinuities into flow discontinuities in rigid scenes, and these discontinuities often align with object boundaries~\cite{longuet1980movingretinal,horn1981opticalflow,brox2010trajectories,ochs2013moving}. This link allows us to convert raw video into object-level supervision without human annotations on the target videos or camera calibration.

We formalize this intuition under a pinhole camera model. Let $p=(u,v)$ denote an image pixel, $D(u,v)$ denote its depth, and $K$ denote the intrinsic matrix. The pixel corresponds to a 3D point $P = D(u,v)K^{-1}[u,v,1]^\top$. Under a rigid camera motion $(R,t) \in SE(3)$, this point projects to a new pixel $p'=(u',v')$ in the next frame:
\begin{equation}
\begin{bmatrix}
u' \\
v' \\
1
\end{bmatrix}
\sim K\left(RD(u,v)K^{-1}
\begin{bmatrix}
u \\
v \\
1
\end{bmatrix}+t\right).
\label{eq:rigid_projection}
\end{equation}
The optical flow at $p$ is therefore a function of pixel location, depth, camera motion, and camera intrinsics:
\begin{equation}
\mathbf{w}(p)=
\begin{bmatrix}
u'-u \\
v'-v
\end{bmatrix}
=\phi(p,D(p);R,t,K).
\label{eq:flow_depth_function}
\end{equation}
Let $p_0$ lie on a depth boundary separating two regions $\Omega^-$ and $\Omega^+$. We use $D^\pm$ and $\mathbf{w}^\pm$ to denote the one-sided depth and optical-flow limits, respectively, as $p$ approaches $p_0$ from within $\Omega^\pm$. Under the same rigid camera motion on both sides, the flow limits and their jump are
\begin{equation}
\begin{aligned}
[\mathbf{w}]_{p_0}
&= \phi(p_0,D^+;R,t,K)
-\phi(p_0,D^-;R,t,K).
\end{aligned}
\label{eq:flow_jump}
\end{equation}
When the camera motion contains a non-degenerate translational component, the projected displacement generally depends on depth; consequently, $D^+\neq D^-$ can induce $\mathbf{w}^+\neq\mathbf{w}^-$. Independently moving objects provide another source of flow discontinuities because their motion differs from that of the surrounding scene. Motion boundaries therefore provide a practical, though not one-to-one, proxy for object boundaries, while the optical flow field already encodes the required geometric information.

This interpretation is consistent with Marr's classical account of motion boundaries~\cite{marr1982vision}:

\begin{quote}
\itshape
``...the velocity field of motion in the image varies continuously almost everywhere, and if it is ever discontinuous at more than an isolated point, then a failure of rigidity (like an object boundary) is present in the outside world. In particular, if the direction of motion is ever discontinuous at more than one point---along a line, for example---then an object boundary is present.''
\end{quote}

We convert this geometric cue into pseudo-instance masks by grouping pixels whose flow vectors form coherent regions.

\subsection{Cycle-1: Pseudo-label Generation}
\label{sec:data_processing}

\textbf{Optical flow estimation.} We process all sources with the same video-to-flow pipeline. We decode videos into short clips and run VideoFlow~\cite{shi2023videoflow}, a multi-frame flow estimator, around each target frame. Each clip contains five frames, and VideoFlow estimates optical flow for the middle frames. We then compute a valid-flow mask using a forward-backward consistency check. Let $\mathbf{w}_f=\mathbf{w}_{t\rightarrow t+1}$ denote the forward flow and $\mathbf{w}_b=\mathbf{w}_{t+1\rightarrow t}$ denote the backward flow. A pixel is considered valid when the forward-backward residual is smaller than a flow-magnitude-aware threshold:
\begin{equation}
\begin{gathered}
p' = p + \mathbf{w}_f(p),
\qquad
r(p) = \bigl\lVert \mathbf{w}_f(p)+\mathbf{w}_b(p') \bigr\rVert_2, \\
\tau(p) = \alpha\!\left(
  \bigl\lVert \mathbf{w}_f(p) \bigr\rVert_2
  + \bigl\lVert \mathbf{w}_b(p') \bigr\rVert_2
\right) + \beta, \\
M(p) = \mathbf{1}\!\left\{r(p)<\tau(p)\right\}.
\end{gathered}
\label{eq:fb_consistency}
\end{equation}
where $M(p)$ is the valid-flow mask and $(\alpha,\beta)$ are tolerance parameters. For corpus-scale storage, we encode each flow field and its valid mask as a compact 16-bit RGB PNG: the first two channels store linearly quantized horizontal and vertical flow values after clipping to a fixed range, and the third channel stores the valid-flow mask.

\textbf{Pseudo-mask construction and filtering.} We construct pseudo-instance masks from the stored flow fields and valid-flow masks. Starting from each unvisited valid pixel, breadth-first search traverses its 4-connected neighbors and assigns a neighboring pixel to the same cluster when the Euclidean difference between their flow vectors is not greater than $\theta_f=1.5$. We discard clusters containing fewer than $\theta_s=100$ pixels. The complete procedure is provided in Appendix~A of the separate appendices file. Figure~\ref{fig:pseudo_label_examples} shows representative input frames, their optical flow, and the resulting pseudo-labels. We then apply several quality filters to remove unreliable pseudo-labels. We retain only frames whose forward-backward valid-flow ratio exceeds 85\%. Because we treat the largest pseudo-label as background, we require it to form a horizontally spanning connected region. Frames that do not satisfy this condition are discarded, reducing the risk that an erroneous flow cluster is used as the background label. Finally, we retain frames whose pseudo-masks contain at least one foreground region. After processing and filtering, Cycle-1 yields 195M pseudo-labeled training frames, compared with 48M frames in the conference version.

\begin{figure}[t]
  \centering
  \setlength{\tabcolsep}{1pt}
  \begin{tabular}{@{}ccc@{}}
    \footnotesize Input & \footnotesize Flow & \footnotesize Label \\[1pt]
    \includegraphics[width=0.31\columnwidth]{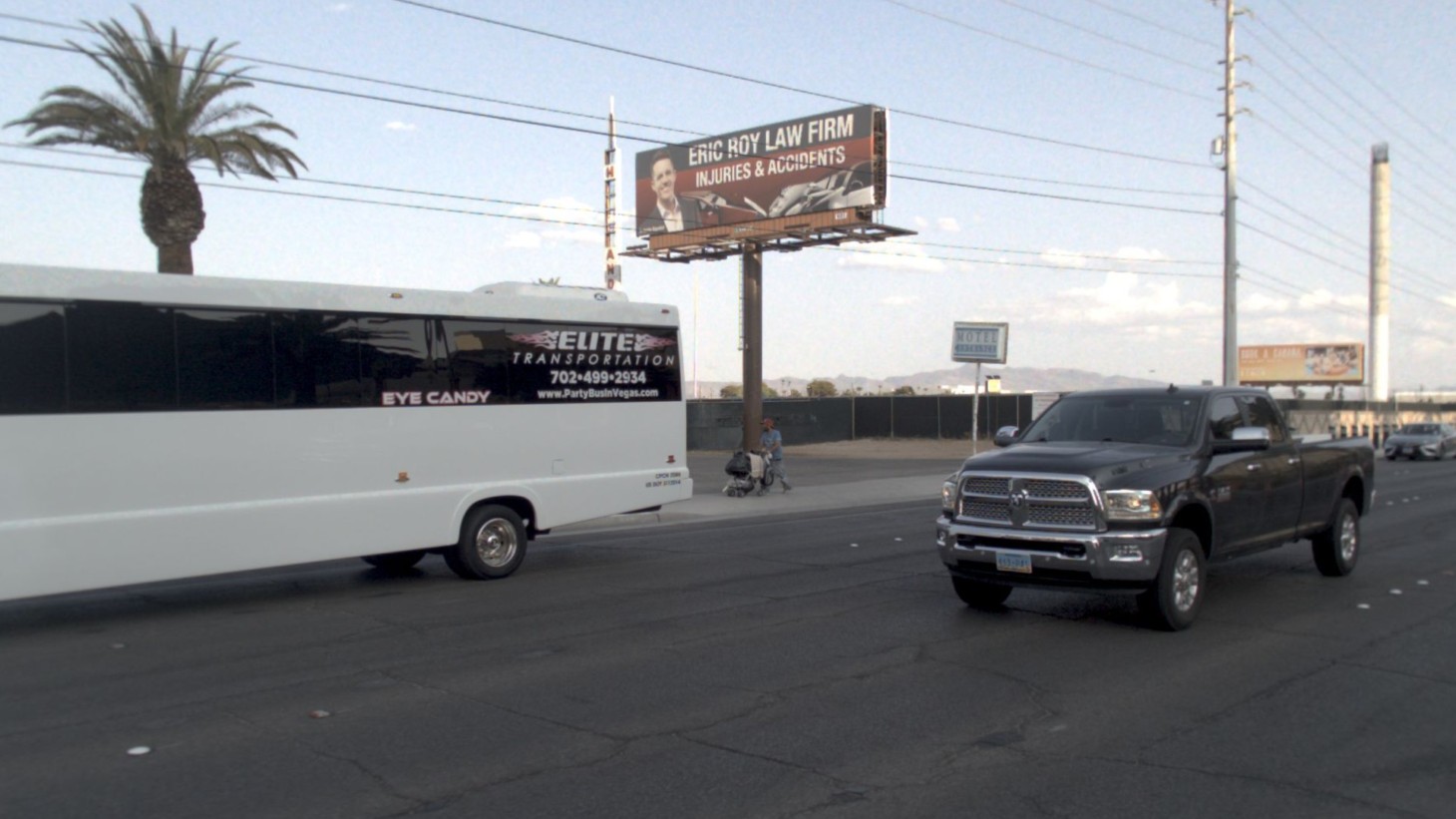} &
    \includegraphics[width=0.31\columnwidth]{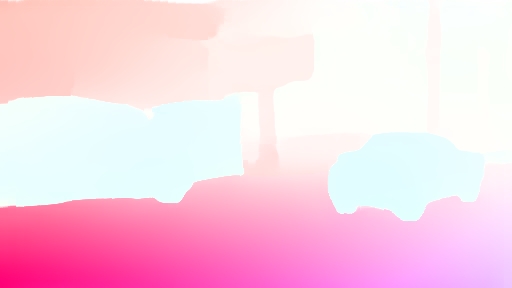} &
    \includegraphics[width=0.31\columnwidth]{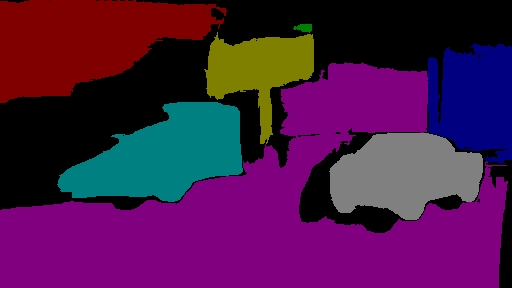} \\[1pt]
    \includegraphics[width=0.31\columnwidth]{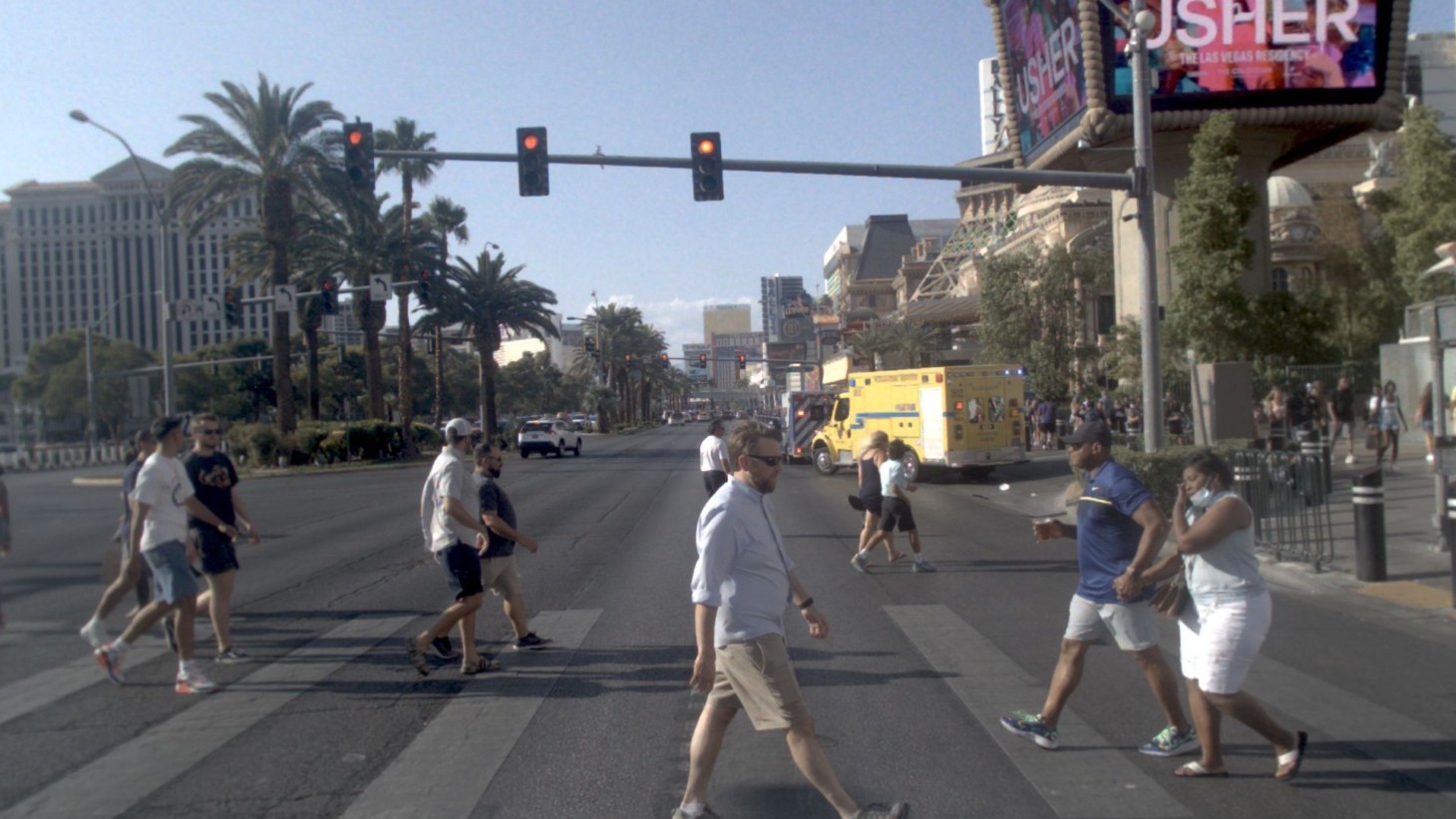} &
    \includegraphics[width=0.31\columnwidth]{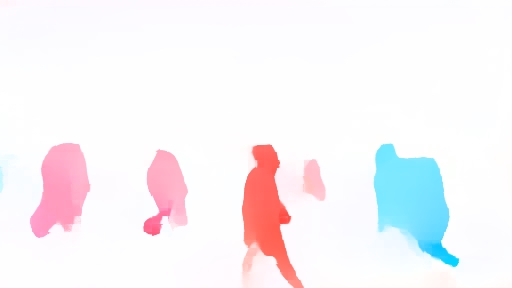} &
    \includegraphics[width=0.31\columnwidth]{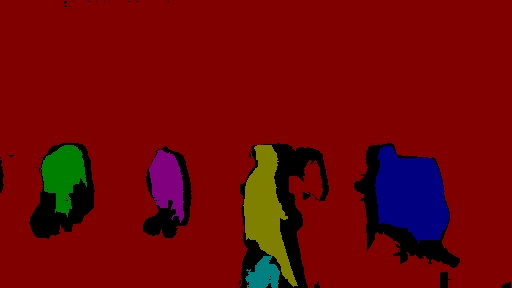} \\[1pt]
    \includegraphics[width=0.31\columnwidth]{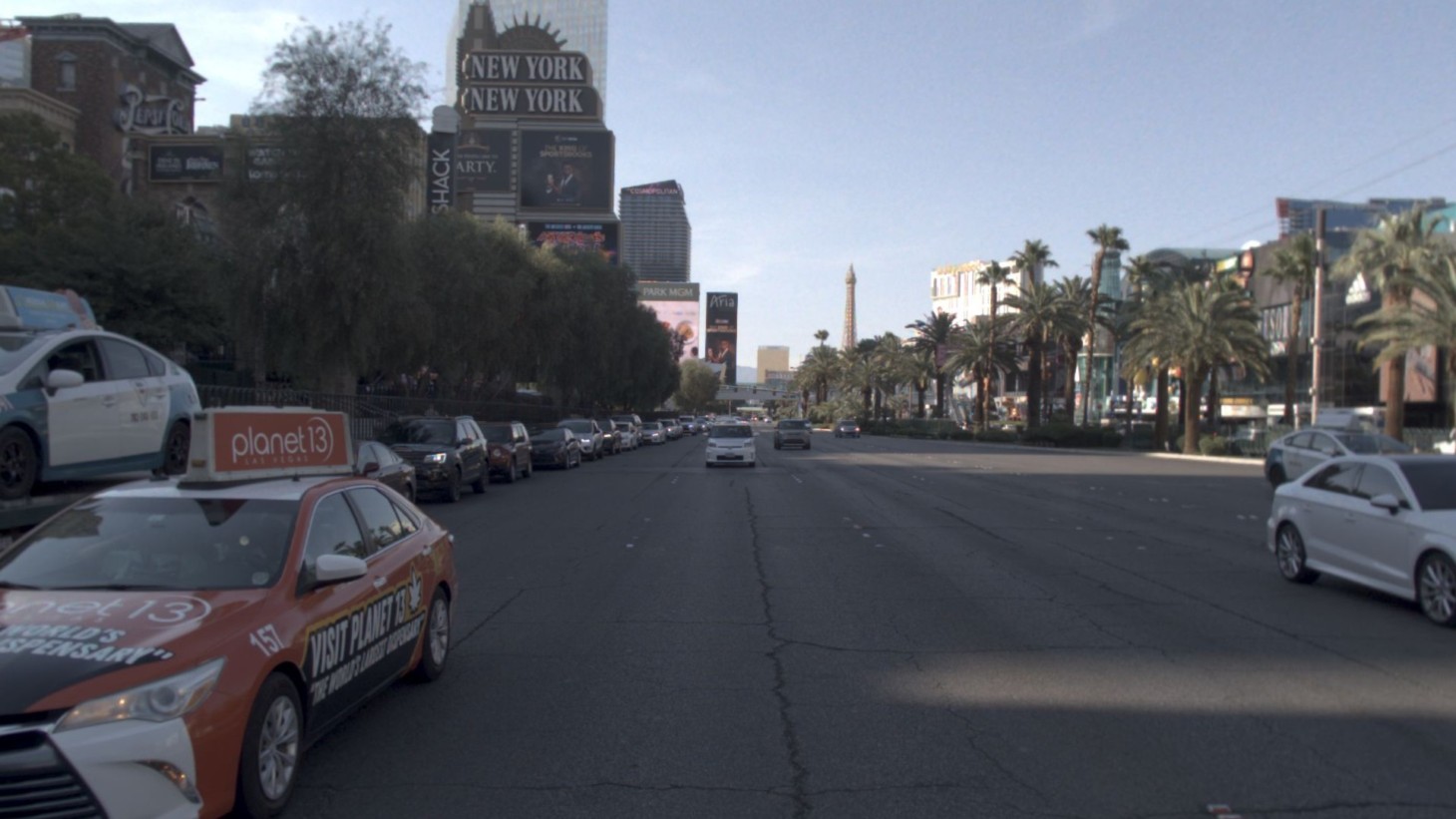} &
    \includegraphics[width=0.31\columnwidth]{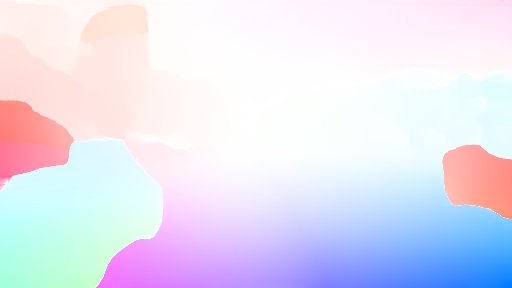} &
    \includegraphics[width=0.31\columnwidth]{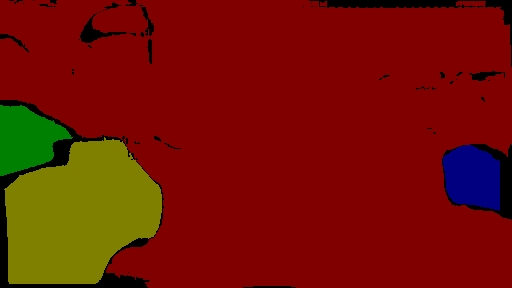}
  \end{tabular}
  \caption{Examples of Cycle-1 pseudo-label generation. We estimate optical flow from neighboring frames and cluster reliable pixels with coherent motion into category-agnostic pseudo-instance labels. Colors distinguish motion clusters, not semantic categories.}
  \label{fig:pseudo_label_examples}
\end{figure}

\subsection{Cycle-2: Motion-Verified Self-Training}
\label{sec:motion_verified_self_training}
The Cycle-1 label-generation pipeline uses motion cues in a conservative way: a pseudo-instance is retained only when the flow field supports a sufficiently complete and reliable motion-induced region. In practice, many frames contain useful but incomplete motion evidence. For example, an object boundary may be partially missing, locally noisy, or connected to nearby regions, preventing BFS clustering from producing a clean closed mask. Such frames may fail to produce complete pseudo-instance masks in Cycle-1, even though their local motion cues still provide valuable information about object separation. Motion-Verified Self-Training is designed to recover this unused supervision by using the Cycle-1 encoder to propose complete object masks, while motion cues are reused to verify and refine these proposals.

\begin{figure*}[t]
\centering
\includegraphics[width=0.98\textwidth]{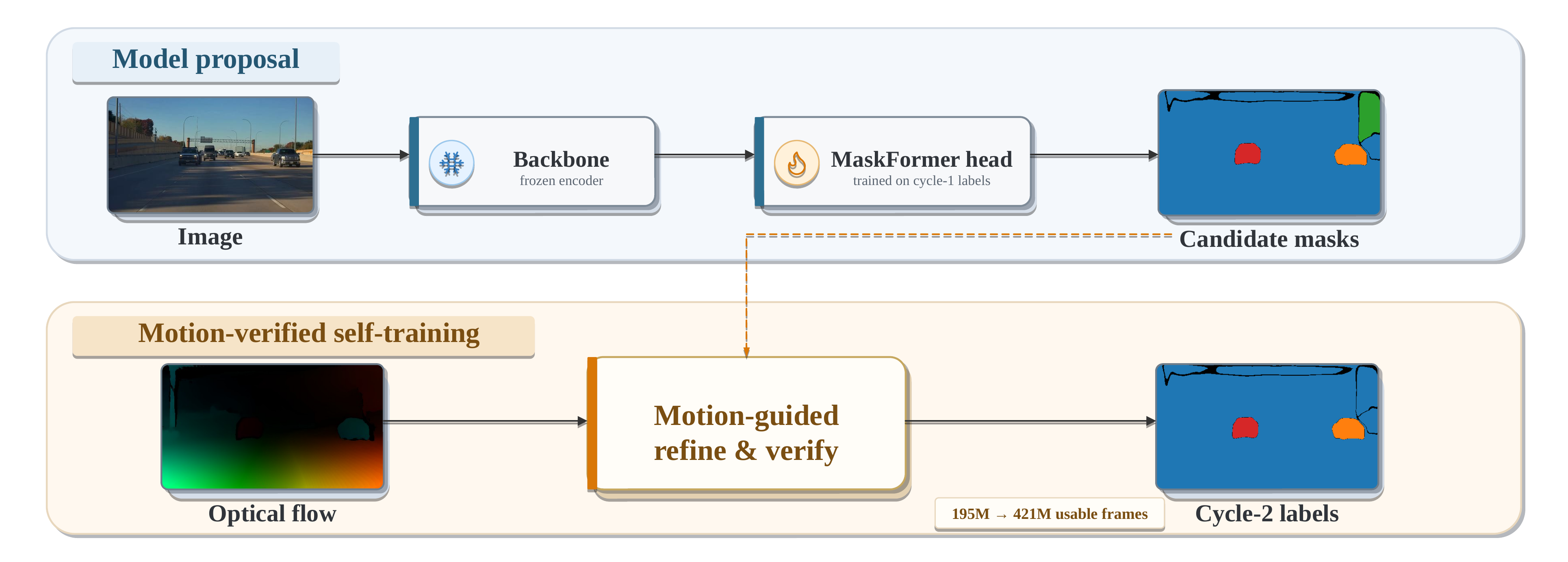}
\caption{Motion-Verified Self-Training. We freeze the Cycle-1 Swin-H encoder and train a MaskFormer head on Cycle-1 pseudo-labels. The resulting proposal model generates candidate masks, which are not accepted directly: optical flow is reused as an independent signal to refine and verify them, producing Cycle-2 pseudo-labels and expanding the training set from 195M to 421M pseudo-labeled frames on the same video sources.}
\label{fig:motion_verified_self_training}
\end{figure*}

\noindent
\textbf{Proposal generation.}
Figure~\ref{fig:motion_verified_self_training} summarizes Cycle-2. We use the Cycle-1 Swin-H encoder pretrained on 195M motion-derived pseudo-labeled frames. With the encoder frozen, we train a MaskFormer head~\cite{cheng2021maskformer} from its multiscale features using the Cycle-1 motion-derived pseudo-labels as supervision. Together, they form a MaskFormer proposal model that predicts candidate masks from image appearance. We apply this model to the same video sources described in Sec.~\ref{sec:exp_implementation}, including frames for which conservative motion clustering does not yield complete pseudo-instance masks. Thus, Motion-Verified Self-Training increases label coverage from the original videos rather than introducing additional data sources.

\noindent
\textbf{Motion-guided refinement and verification.}
Inspired by the data-engine design of SAM~2~\cite{ravi2024sam2}, we do not treat self-training as simply accepting the model's own predictions. Instead, we introduce motion as an additional supervisory signal so that model-generated masks are refined and verified before they become new labels. Our procedure extracts a reliable interior seed from each mask proposal, expands the seed using feature similarity while preventing growth across flow-magnitude edges, and retains the refined mask only if it is supported by the flow field. Let $P$ denote a mask proposal and $\mathbf{z}(p)$ the dense learned feature at pixel $p$. We obtain a reliable interior seed by adaptive erosion,
\begin{equation}
S_P = \operatorname{Erode}(P,r_P),
\qquad
\boldsymbol{\mu}_P = \frac{1}{|S_P|}\sum_{p\in S_P}\frac{\mathbf{z}(p)}{\|\mathbf{z}(p)\|_2},
\label{eq:motion_refine_seed}
\end{equation}
where $r_P$ is an adaptive erosion radius whose computation is detailed in Appendix~B of the separate appendices file. Starting from $S_P^{(0)}=S_P$, we grow the seed within a local proposal neighborhood according to feature similarity, while preventing it from crossing strong flow-magnitude edges. With $\mathbf{w}$ denoting optical flow, let $a_w(p)=\|\mathbf{w}(p)\|_2$ be its magnitude. We define the flow-magnitude-edge indicator
\begin{equation}
E_w(p) = \mathbf{1}\left[\|\nabla a_w(p)\|_2 \geq \tau_{\mathrm{edge}}\right].
\label{eq:flow_edge}
\end{equation}
For a candidate pixel, we use cosine similarity to measure its similarity to the mask proposal
\begin{equation}
s_P(p)=\frac{\mathbf{z}(p)^\top\boldsymbol{\mu}_P}{\|\mathbf{z}(p)\|_2\|\boldsymbol{\mu}_P\|_2}.
\label{eq:motion_feature_similarity}
\end{equation}
A pixel is admissible if it is feature-compatible with the proposal and does not lie on a flow-magnitude edge:
\begin{equation}
\mathcal{A}_P=\left\{p:s_P(p)\geq\tau_{\mathrm{feat}},\ E_w(p)=0\right\}.
\label{eq:motion_admissible_set}
\end{equation}
At iteration $k$, the refined region is updated by
\begin{equation}
S_P^{(k+1)}=S_P^{(k)}\cup
\left(\mathcal{N}(S_P^{(k)})\cap\mathcal{A}_P\right),
\label{eq:motion_refine_growth}
\end{equation}
where $\mathcal{N}(\cdot)$ denotes the 8-connected neighborhood and $\tau_{\mathrm{feat}}$ is a fixed feature-similarity threshold. Thus, model features complete object regions from appearance cues, while flow discontinuities constrain the completion.

After refinement, we verify the resulting proposal $\widetilde{P}$ using flow support. A reliable mask should have boundaries supported by flow-magnitude edges and should contain limited magnitude edges in its interior. Let $\partial\widetilde{P}$ be the boundary of the mask. We compute
\begin{equation}
\begin{aligned}
\rho_{\mathrm{bdry}}(\widetilde{P})
&= \frac{|\partial\widetilde{P}\cap E_w|}
{|\partial\widetilde{P}|},\\
\rho_{\mathrm{int}}(\widetilde{P})
&= \frac{|(\widetilde{P}\setminus\partial\widetilde{P})\cap E_w|}
{|\widetilde{P}\setminus\partial\widetilde{P}|}.
\end{aligned}
\label{eq:motion_verification}
\end{equation}
The proposal is accepted only when its boundary has sufficient flow-magnitude-edge support and its interior contains limited flow-magnitude conflict. Masks with weak boundary support or high interior conflict are rejected. Finally, we merge back Cycle-1 motion-derived foreground labels that are not covered by the refined model proposals. This preserves reliable Cycle-1 supervision while allowing model proposals to expand label coverage. The label-generation hyperparameters are summarized in Appendix~B of the separate appendices file.

Figure~\ref{fig:cycle2_rule_qualitative} visualizes the intermediate states of these rules on representative proposals. The first two rows show proposals that pass verification: erosion produces a compact interior seed, and feature-compatible constrained growth respects the flow-magnitude edges. The last row passes the boundary-support criterion but is rejected because of substantial interior flow conflict.

\begin{figure*}[t]
\centering
\setlength{\tabcolsep}{1pt}
\begin{tabular}{@{}ccccc@{}}
{\scriptsize Proposal $P$} &
{\scriptsize Interior seed $S_P$} &
{\scriptsize Constrained growth} &
{\scriptsize Boundary support} &
{\scriptsize Interior conflict} \\[1pt]
\includegraphics[width=0.19\textwidth]{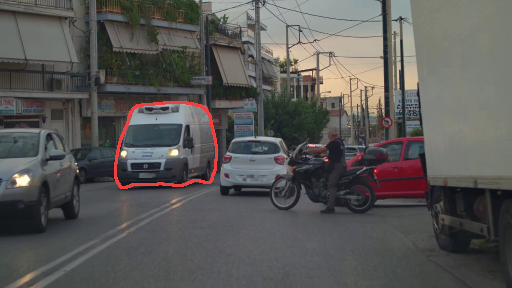} &
\includegraphics[width=0.19\textwidth]{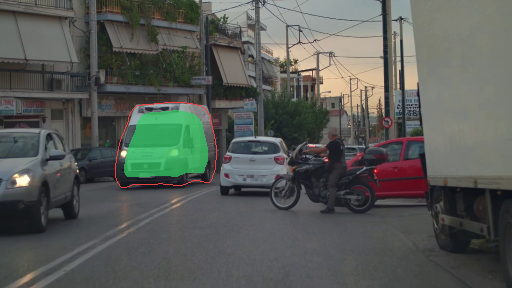} &
\includegraphics[width=0.19\textwidth]{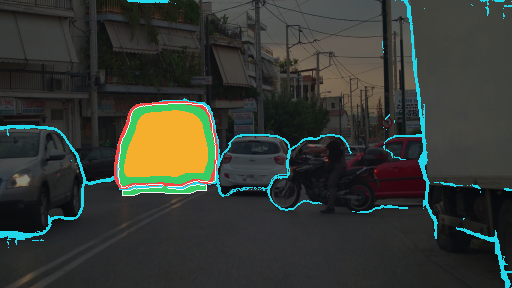} &
\includegraphics[width=0.19\textwidth]{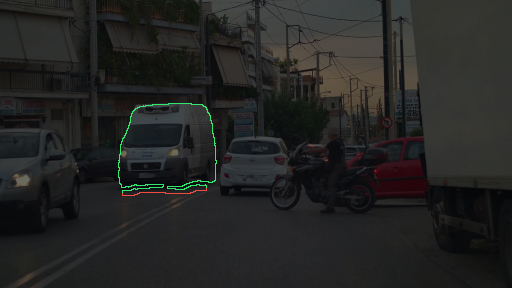} &
\includegraphics[width=0.19\textwidth]{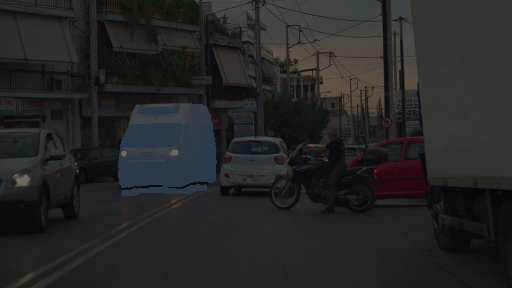} \\
\multicolumn{5}{c}{\scriptsize \textbf{Accepted}\quad $\rho_{\mathrm{bdry}}=0.83,\ \rho_{\mathrm{int}}=0.00$} \\[2pt]
\includegraphics[width=0.19\textwidth]{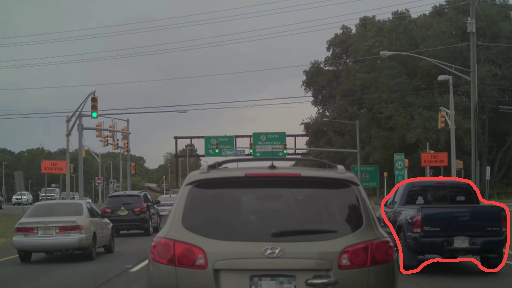} &
\includegraphics[width=0.19\textwidth]{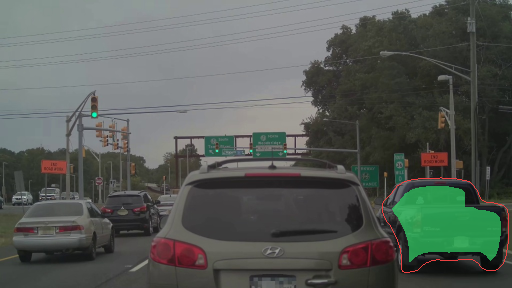} &
\includegraphics[width=0.19\textwidth]{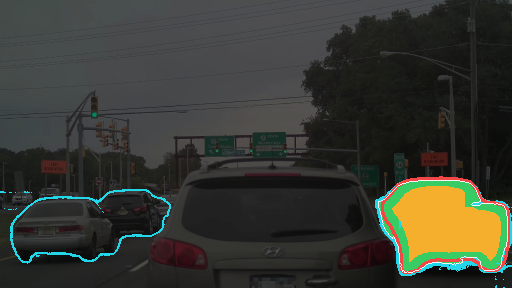} &
\includegraphics[width=0.19\textwidth]{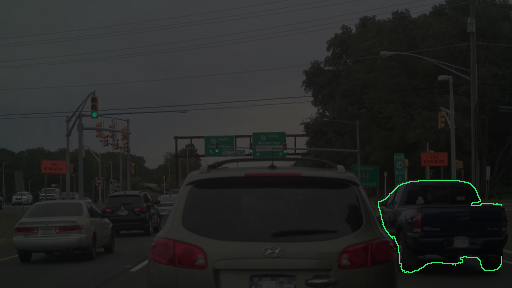} &
\includegraphics[width=0.19\textwidth]{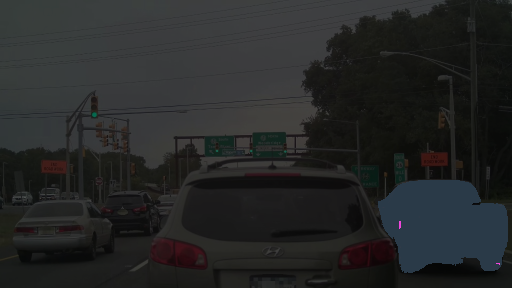} \\
\multicolumn{5}{c}{\scriptsize \textbf{Accepted}\quad $\rho_{\mathrm{bdry}}=1.00,\ \rho_{\mathrm{int}}=0.00$} \\[2pt]
\includegraphics[width=0.19\textwidth]{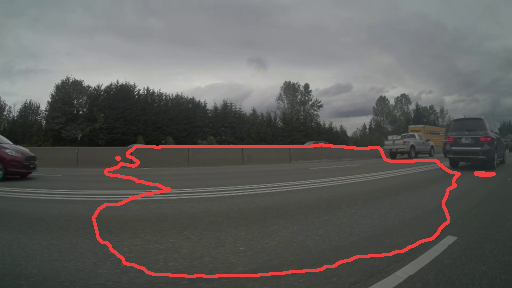} &
\includegraphics[width=0.19\textwidth]{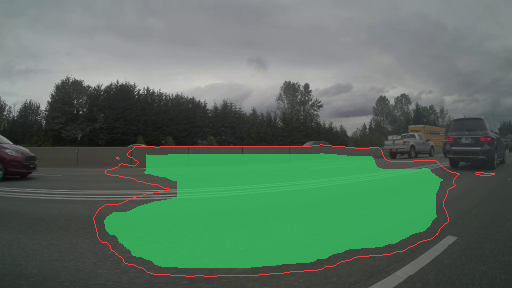} &
\includegraphics[width=0.19\textwidth]{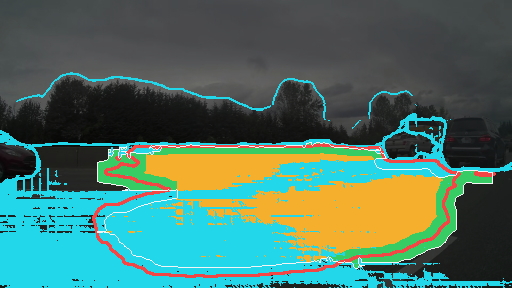} &
\includegraphics[width=0.19\textwidth]{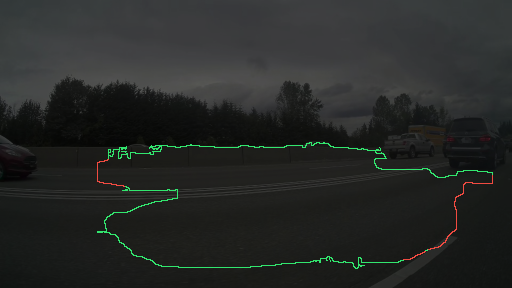} &
\includegraphics[width=0.19\textwidth]{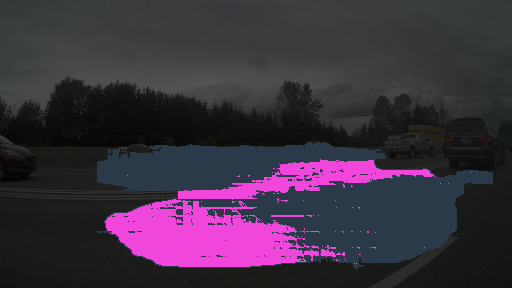} \\
\multicolumn{5}{c}{\scriptsize \textbf{Rejected}\quad $\rho_{\mathrm{bdry}}=0.84,\ \rho_{\mathrm{int}}=0.35$}
\end{tabular}
\caption{Qualitative visualization of the Cycle-2 motion-guided refinement and verification rules. Each row follows a proposal $P$ through adaptive erosion, constrained growth, and flow-based verification. In the growth panel, orange denotes the seed, green denotes added pixels, and cyan denotes flow-magnitude edges. In the boundary-support panel, green and red denote supported and unsupported boundary pixels, respectively; magenta in the interior-conflict panel marks flow edges inside the refined mask. Proposals are accepted when $\rho_{\mathrm{bdry}}\geq0.4$ and $\rho_{\mathrm{int}}\leq0.05$.}
\label{fig:cycle2_rule_qualitative}
\end{figure*}

After motion-guided refinement and verification, Motion-Verified Self-Training expands the training set from 195M Cycle-1 pseudo-labeled frames to 421M Cycle-2 pseudo-labeled frames on the same video sources. We train a Cycle-2 Swin-H teacher on this expanded set and subsequently distill it into the deployed backbone family. Figure~\ref{fig:cycle_label_coverage} shows representative label expansion; the complete pipeline effect, coverage, and agreement are analyzed in Sec.~\ref{sec:exp_mvst_effect}.

\subsection{Training Details}
\label{sec:training_details}
\textbf{Network architecture.}
The representation network takes a single image as input and consists of a backbone, a Feature Pyramid Network (FPN), and a dense prediction head based on the semantic branch of Panoptic FPN~\cite{kirillov2019panoptic}. The FPN fuses multiscale backbone features, and the prediction head produces a 64-channel dense feature map. We $\ell_2$-normalize each sampled pixel feature before applying the pairwise metric-learning objective.

\noindent
\textbf{Pixel-level objective.}
The pseudo-instance masks supervise the network through a pixel-level metric-learning objective. For each augmented training crop, we sample 200 pixels from the valid-flow regions defined by the forward-backward consistency check in Sec.~\ref{sec:data_processing}, distributing them approximately evenly across the pseudo-labels present in the crop. We form all within-image pairs among these sampled pixels and optimize their feature distances according to the corresponding pseudo-labels. Pixels from the same foreground cluster move closer in feature space, while pixels from different clusters move apart by a fixed margin. We ignore pairs where both pixels belong to the background, because the pseudo-labels cover only a subset of object instances and the remaining background may contain unlabeled pixels that belong to different objects.

Formally, let $\mathbf{z}_i$ and $\mathbf{z}_j$ denote the normalized learned feature vectors at sampled pixels $i$ and $j$, and let $y_i$ and $y_j$ denote their pseudo-instance labels. We treat the pseudo-label with the largest area in each frame as background and assign it label $0$. We define the pairwise objective as
\begin{equation}
\mathcal{L}(i,j)=
\begin{cases}
\lVert \mathbf{z}_i - \mathbf{z}_j \rVert_2^2, & y_i = y_j \ne 0,\\
\max(m - \lVert \mathbf{z}_i - \mathbf{z}_j \rVert_2, 0)^2, & y_i \ne y_j,\\
0, & y_i = y_j = 0,
\end{cases}
\label{eq:pairwise_loss}
\end{equation}
where the margin is set to $m=1.0$. Let $\mathcal{P}$ denote all pairs formed from the 200 sampled pixels after excluding background-background pairs. The total training loss is
\begin{equation}
\mathcal{L}_{\mathrm{total}} = \frac{1}{\lvert\mathcal{P}\rvert}
\sum_{(i,j)\in\mathcal{P}}\mathcal{L}(i,j).
\label{eq:total_loss}
\end{equation}

This objective encourages the encoder to capture object unity and instance separation without explicitly supervising category identity. %Although video clips provide the supervision during pretraining, the encoder itself takes only a single image as input. After pretraining, it can therefore serve as an ordinary image backbone without optical flow.
It learns appearance cues for recovering object-level grouping from a static image.

\section{Experiments}
\label{sec:experiments}

\subsection{Implementation Details}
\label{sec:exp_implementation}
\textbf{Data sources.} We use four video sources for pretraining. OpenDV-YouTube~\cite{yang2024genad} contains front-view driving videos collected from more than 244 cities, with about 1,747 hours of raw video. nuPlan~\cite{caesar2021nuplan} provides 1,200 hours of driving logs from four cities, including 120 hours with eight camera views. We further add 1,700 hours from NVIDIA PhysicalAI-Autonomous-Vehicles~\cite{nvidia2025physicalai} and 2,516 hours of self-collected web videos. In total, the pretraining corpus contains 7,163 hours of raw video. The same video sources are used in both training cycles; Cycle-2 increases the number of usable pseudo-labeled frames rather than introducing new sources.

\noindent
\textbf{Cycle-1 representation pretraining.}
We train a Swin-H encoder from scratch on the 195M motion-derived pseudo-labeled frames for 50 epochs. Training uses $224\times224$ crops, a global batch size of 16,384, and the Muon optimizer~\cite{jordan2024muon} with linear warmup followed by cosine decay. The encoder is trained with the pixel-level pairwise objective described in Sec.~\ref{sec:training_details}. The Cycle-1 pretraining run takes approximately 374 hours of wall-clock time on 128 NVIDIA H20 GPUs.

\noindent
\textbf{Cycle-2 label generation and training.}
Starting from the Cycle-1 checkpoint, we freeze the Swin-H encoder and train a class-agnostic MaskFormer head~\cite{cheng2021maskformer} to generate candidate object masks. After motion-guided refinement and verification, we initialize a second Swin-H encoder from the Cycle-1 checkpoint and fine-tune it on the resulting 421M pseudo-labeled frames using the same representation objective and augmentations. This Swin-H encoder, trained directly on the Cycle-2 pseudo-labeled frames, serves as the teacher for the subsequent representation-distillation stage.

\noindent
\textbf{Representation distillation.}
Following the teacher--student knowledge-distillation paradigm~\cite{hinton2015distilling}, we use the Cycle-2 Swin-H encoder as a frozen teacher and train newly initialized Swin-T/S/B/L students, together with a Swin-H student with the same architecture, to match its dense representation. The distillation objective minimizes the pixel-wise cosine distance between normalized teacher and student features. For Swin-T/S/B/L, this transfers the representation to smaller architectures. For Swin-H, same-architecture feature self-distillation is a post-Cycle-2 regularizing retraining stage rather than architectural compression: the student matches continuous teacher features instead of directly optimizing against pseudo-instance labels. This may reduce sensitivity to pseudo-label noise, consistent with the observation that self-distillation can improve upon the teacher~\cite{furlanello2018born}. Only the student backbone is retained for downstream transfer.

\noindent
\textbf{Downstream transfer.}
Unless otherwise specified, \emph{Ours} denotes the final Cycle-2 distilled model family: all scales from Swin-T to Swin-H are students distilled from the Cycle-2 Swin-H teacher. Thus, the Swin-H results reported below use the same-architecture self-distilled student rather than the teacher. For the supervised baseline, we use the official ImageNet-22K-pretrained checkpoints for Swin-T/S/B/L~\cite{liu2021swin}. For all downstream tasks, only the pretrained backbone is transferred; task-specific components are initialized and trained separately using the corresponding official codebase.

\subsection{Qualitative Results}
\label{sec:exp_qualitative}
As already suggested by the feature-map comparison in Fig.~1, our motion-derived representation organizes visual evidence at the object-instance level rather than at the category or texture level. To inspect this behavior more locally, we compute reference-point similarity maps by measuring the cosine similarity between each colored reference feature and all spatial features in the same image.

\begin{figure*}[t]
\centering
\includegraphics[width=0.98\textwidth]{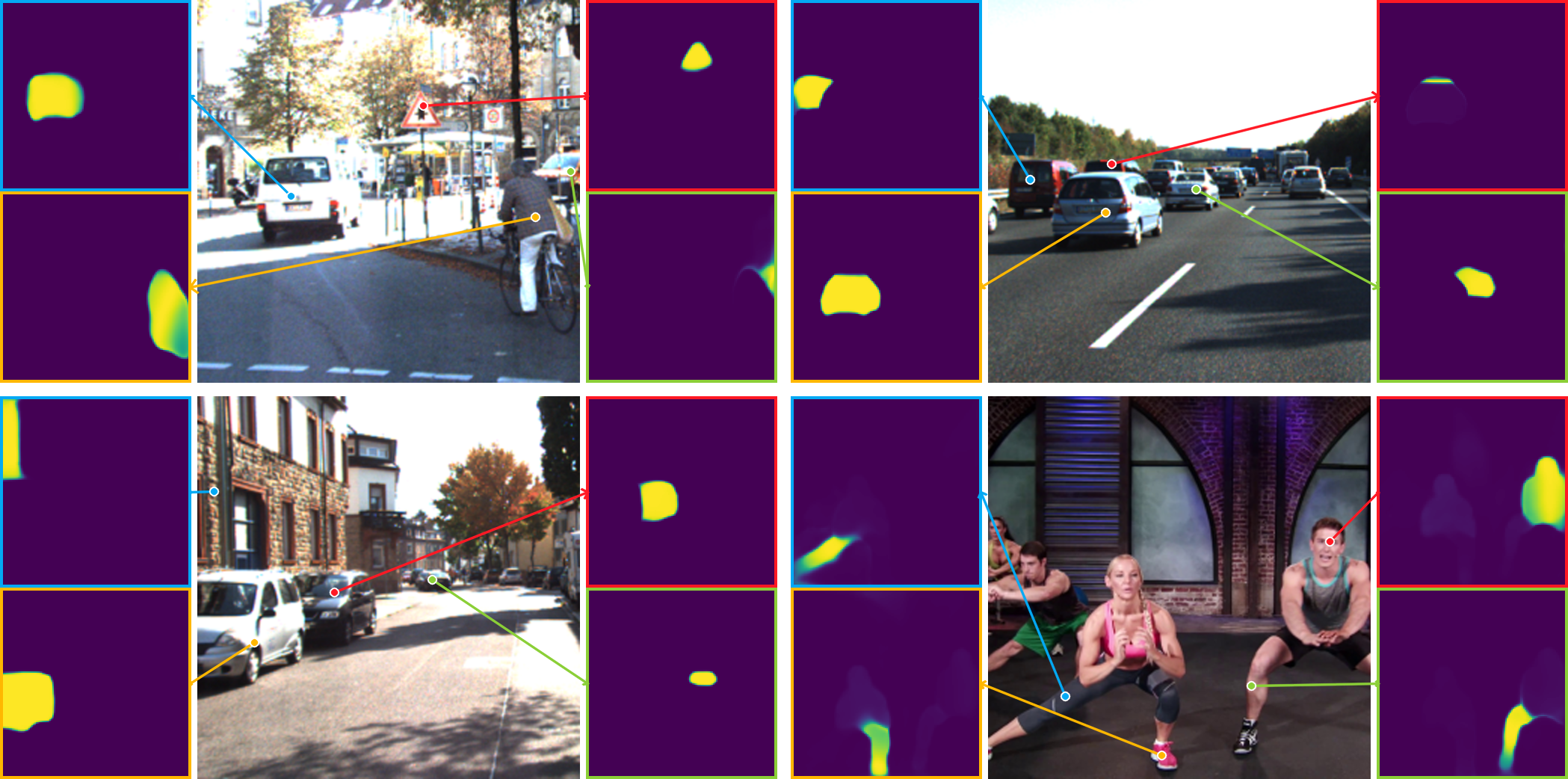}
\caption{Reference-point similarity maps of our Swin-H encoder. Each panel places four colored reference points on the input image and visualizes, around the image, the cosine-similarity map induced by each point. High responses remain localized around coherent object regions, indicating that the learned features encode instance-level affinity rather than only category-level appearance.}
\label{fig:qual_similarity}
\end{figure*}

Figure~\ref{fig:qual_similarity} shows that the responses are concentrated on coherent object regions in these scenes, instead of spreading broadly across pixels with similar appearance. This mirrors the qualitative pattern in Fig.~1: generic foundation models such as DINOv3, CLIP, and MAE tend to emphasize category-level or texture-level structure, whereas our representation better preserves the unity of each object instance.

\subsection{Monocular Depth Estimation}
\label{sec:exp_depth}
We first evaluate whether the learned representation transfers to low-level geometric prediction. We adopt DCDepth~\cite{NEURIPS2024_76bea0a1} as the depth decoder and evaluate on the KITTI Eigen split~\cite{geiger2012we,eigen2014depth}. We compare our models with the supervised ImageNet-22K-pretrained model, Semantic-SAM~\cite{li2024segment}, SimMIM~\cite{xie2022simmim}, DINOv2~\cite{oquab2023dinov2}, and DINOv3~\cite{simeoni2025dinov3}. Table~\ref{tab:depth_eigen} reports Abs Rel, RMSE, and $\delta_1$ across backbone scales.

As shown in Table~\ref{tab:depth_eigen}, our Swin-L outperforms DINOv3 ViT-L in Abs Rel (0.042 vs.\ 0.044), RMSE (1.715 vs.\ 1.787), and $\delta_1$ (0.988 vs.\ 0.986). Notably, our Swin-L achieves performance close to our Swin-H on depth estimation, with only limited additional gain from scaling to Swin-H.

More broadly, our models achieve the best available results at their respective scales across all three reported metrics. In particular, our Swin-S and Swin-B outperform the corresponding DINOv3 ViT-S and ViT-B baselines in Abs Rel, RMSE, and $\delta_1$.

\begin{table*}[t]
\centering
\caption{Monocular depth estimation on the KITTI Eigen split using DCDepth~\cite{NEURIPS2024_76bea0a1}. Rows denote pretraining methods and columns denote backbone variants. In the shared S/B/L groups, DINO uses ViT backbones, while the other methods use Swin backbones.}
\label{tab:depth_eigen}
\setlength{\tabcolsep}{1.0pt}
\scriptsize
\begin{tabular}{@{}l*{15}{>{\centering\arraybackslash}p{3.9em}}@{}}
\hline
Pretraining & \multicolumn{3}{c}{Swin-T} & \multicolumn{3}{c}{ViT-S/Swin-S} & \multicolumn{3}{c}{ViT-B/Swin-B} & \multicolumn{3}{c}{ViT-L/Swin-L} & \multicolumn{3}{c}{Swin-H} \\
& \mbox{Abs\,Rel$\downarrow$} & RMSE $\downarrow$ & $\delta_1\uparrow$ & \mbox{Abs\,Rel$\downarrow$} & RMSE $\downarrow$ & $\delta_1\uparrow$ & \mbox{Abs\,Rel$\downarrow$} & RMSE $\downarrow$ & $\delta_1\uparrow$ & \mbox{Abs\,Rel$\downarrow$} & RMSE $\downarrow$ & $\delta_1\uparrow$ & \mbox{Abs\,Rel$\downarrow$} & RMSE $\downarrow$ & $\delta_1\uparrow$ \\
\hline
ImageNet-22K & 0.055 & 2.182 & 0.969 & 0.052 & 2.093 & 0.973 & 0.052 & 2.088 & 0.975 & 0.051 & 2.044 & 0.977 & -- & -- & -- \\
SimMIM~\cite{xie2022simmim} & -- & -- & -- & -- & -- & -- & 0.050 & 2.031 & 0.975 & 0.048 & 1.941 & 0.979 & -- & -- & -- \\
Semantic-SAM~\cite{li2024segment} & 0.055 & 2.169 & 0.971 & -- & -- & -- & -- & -- & -- & 0.049 & 2.007 & 0.979 & -- & -- & -- \\
DINOv2~\cite{oquab2023dinov2} & -- & -- & -- & 0.052 & 2.153 & 0.974 & 0.050 & 2.081 & 0.976 & 0.048 & 1.971 & 0.980 & -- & -- & -- \\
DINOv3~\cite{simeoni2025dinov3} & -- & -- & -- & 0.052 & 2.161 & 0.974 & 0.047 & 1.958 & 0.981 & 0.044 & 1.787 & 0.986 & -- & -- & -- \\
\hline
Ours & \textbf{0.049} & \textbf{1.958} & \textbf{0.978} & \textbf{0.044} & \textbf{1.803} & \textbf{0.985} & \textbf{0.043} & \textbf{1.766} & \textbf{0.986} & \textbf{0.042} & \textbf{1.715} & \textbf{0.988} & 0.043 & 1.712 & 0.988 \\
\hline
\end{tabular}
\end{table*}

We further evaluate on the official KITTI online benchmark. Table~\ref{tab:depth_official} compares our Swin-L and Swin-H models with representative published methods under the official evaluation protocol. Our Swin-H achieves the best performance among our models, obtaining 6.36 Abs Rel, 1.02 Sq Rel, 8.53 iRMSE, and 7.91 SILog.

\begin{table}[t]
\centering
\caption{Monocular depth estimation on the official KITTI online benchmark. Metrics follow the leaderboard convention.}
\label{tab:depth_official}
\setlength{\tabcolsep}{1.5pt}
\scriptsize
\begin{tabular}{llcccc}
\hline
Method & Pretraining & Abs Rel $\downarrow$ & Sq Rel $\downarrow$ & iRMSE $\downarrow$ & SILog $\downarrow$ \\
\hline
NeW CRFs~\cite{yuan2022neural} & ImageNet sup. & 8.37 & 1.83 & 11.03 & 10.39 \\
VA-DepthNet~\cite{liu2023va} & ImageNet sup. & 7.96 & 1.66 & 10.44 & 9.63 \\
IEBins~\cite{shao2023iebins} & MIM~\cite{xie2023revealing} & 7.82 & 1.60 & 10.68 & 9.84 \\
NDDepth~\cite{shao2023nddepth} & MIM~\cite{xie2023revealing} & 7.75 & 1.59 & 10.62 & 9.62 \\
DCDepth~\cite{NEURIPS2024_76bea0a1} & Semantic-SAM~\cite{li2024segment} & 7.83 & 1.54 & 10.12 & 9.60 \\
\hline
DCDepth & Ours (Swin-L) & 6.44 & 1.07 & 8.66 & 8.07 \\
DCDepth & Ours (Swin-H) & \textbf{6.36} & \textbf{1.02} & \textbf{8.53} & \textbf{7.91} \\
\hline
\end{tabular}
\end{table}

\begin{figure*}[t]
\centering
\setlength{\tabcolsep}{2pt}
\begin{tabular}{@{}>{\centering\arraybackslash}m{0.315\textwidth}>{\centering\arraybackslash}m{0.315\textwidth}>{\centering\arraybackslash}m{0.315\textwidth}@{}}
\footnotesize Input & \footnotesize DINOv2 ViT-L & \footnotesize Swin-L (Ours) \\[2pt]
\includegraphics[width=\linewidth]{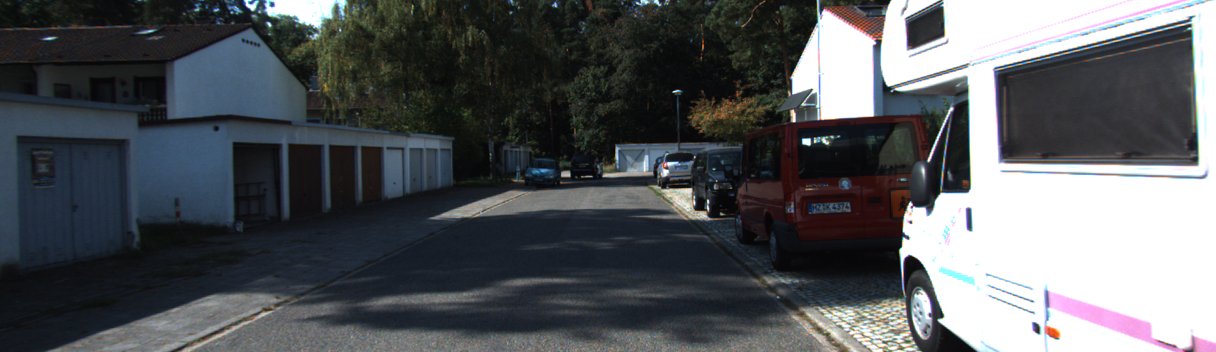} &
\includegraphics[width=\linewidth]{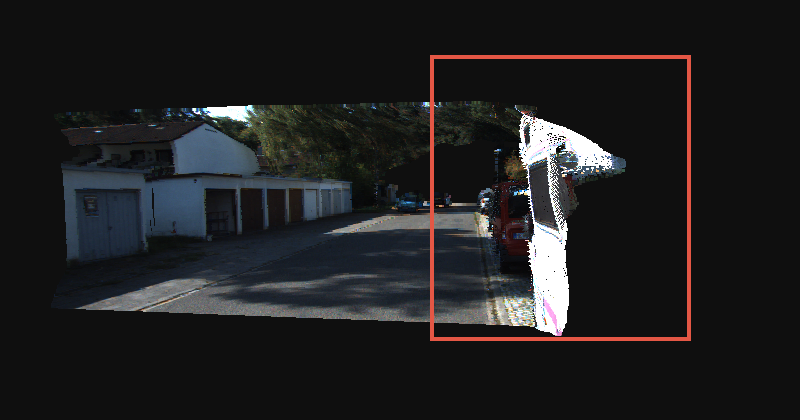} &
\includegraphics[width=\linewidth]{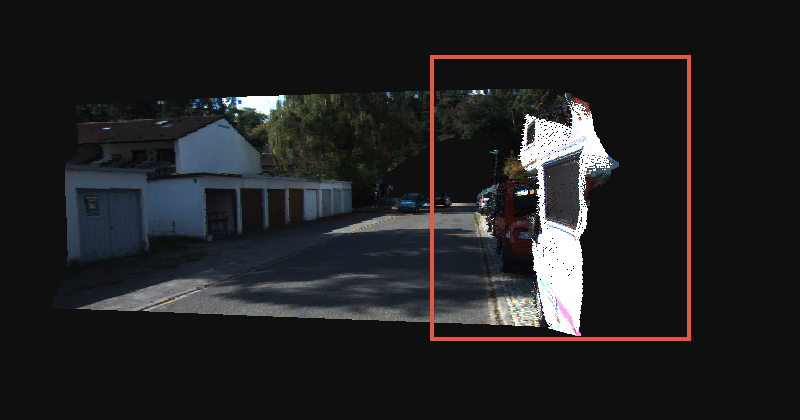} \\[2pt]
\includegraphics[width=\linewidth]{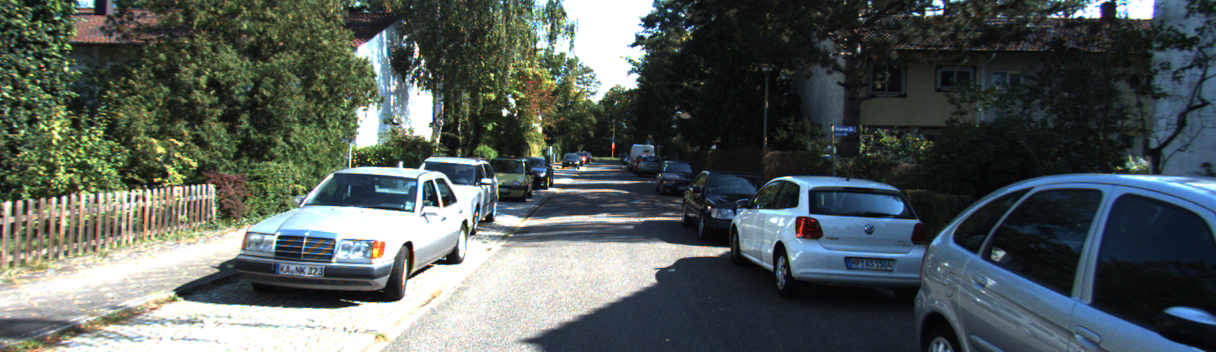} &
\includegraphics[width=\linewidth]{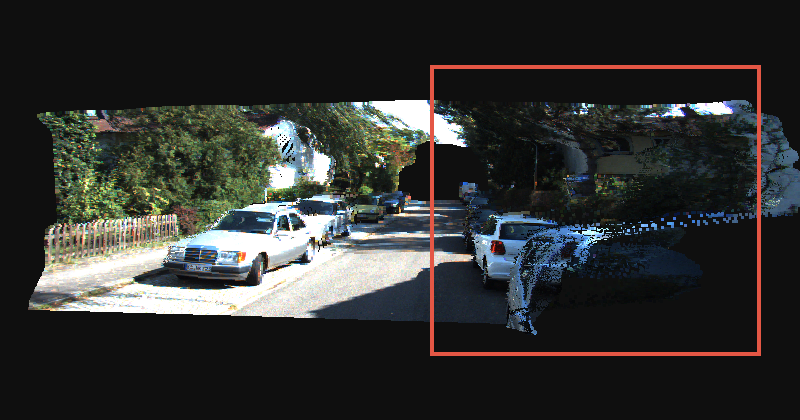} &
\includegraphics[width=\linewidth]{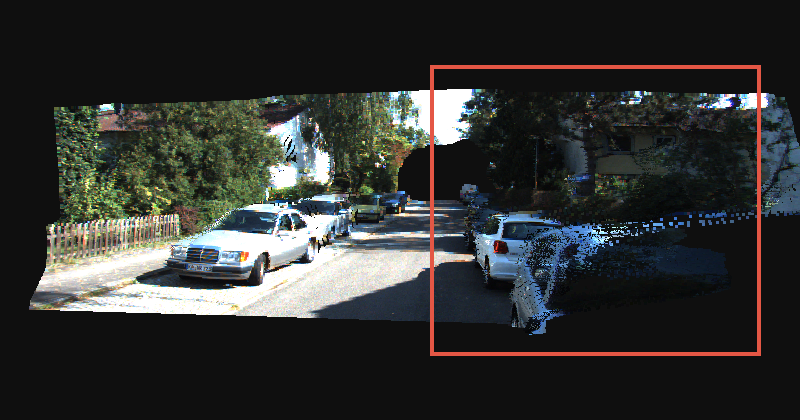} \\[2pt]
\includegraphics[width=\linewidth]{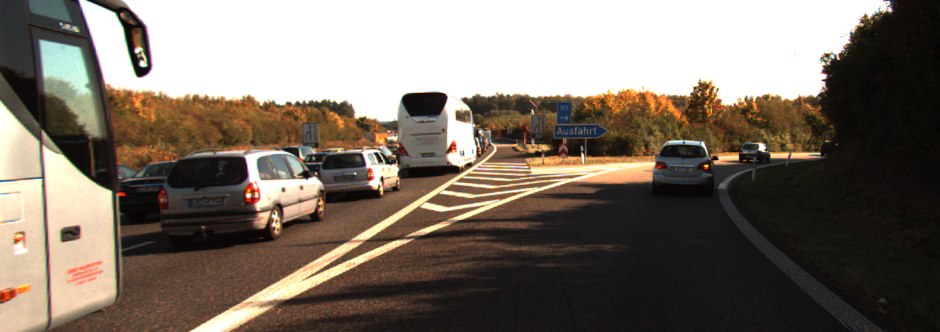} &
\includegraphics[width=\linewidth]{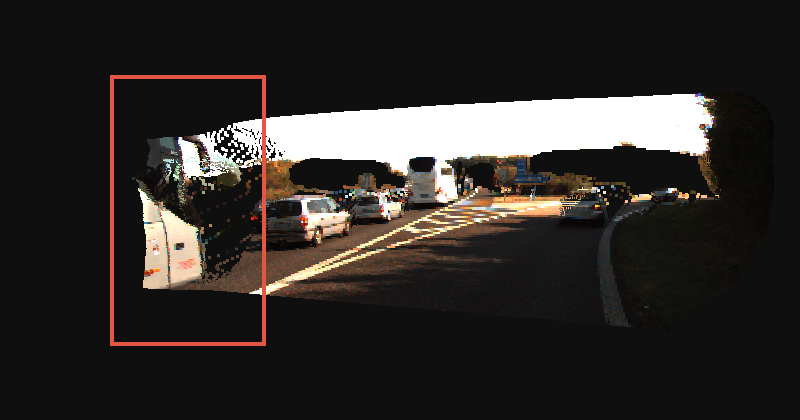} &
\includegraphics[width=\linewidth]{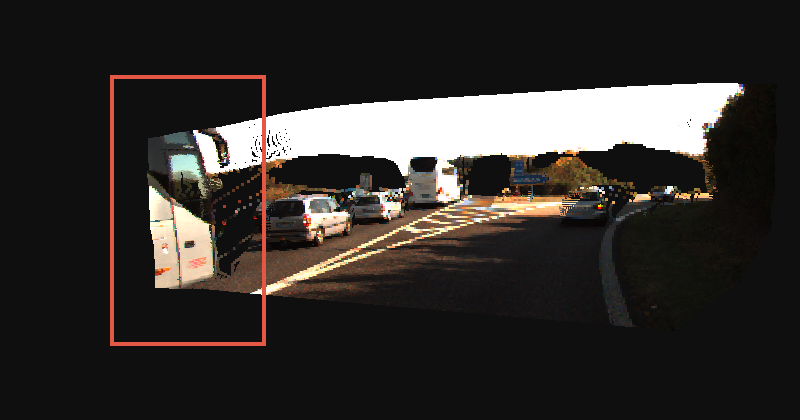}
\end{tabular}
\caption{Qualitative comparison of depth geometry under a 1.5~m lateral novel-view rendering. Our Swin-L produces more coherent surfaces and fewer depth artifacts than DINOv2 ViT-L.}
\label{fig:depth_camera_shift}
\end{figure*}

To inspect the geometric coherence of the predicted depth, we back-project each depth map into an RGB-colored point cloud and render it from a virtual camera translated laterally by 1.5~m (Fig.~\ref{fig:depth_camera_shift}). This rendering makes local depth inconsistencies visible as distorted, fragmented, or spurious surfaces. Compared with DINOv2 ViT-L, our Swin-L better preserves coherent surfaces while reducing foreground--background leakage. This behavior is consistent with our motion-derived supervision, which encourages feature coherence within an object and separation across object boundaries.

\subsection{3D Object Detection}
\label{sec:exp_detection}
We evaluate our learned visual representations on the nuScenes validation set~\cite{caesar2020nuscenes} for 3D object detection, using BEVFormer V2~\cite{yang2023bevformer} as the detection framework. All experiments use two input frames and a 24-epoch downstream schedule. We compare our method with supervised ImageNet-22K pretraining, SimMIM~\cite{xie2022simmim}, DINOv2~\cite{oquab2023dinov2}, and DINOv3~\cite{simeoni2025dinov3}, where results are available.

At the higher input resolution of $1600\times900$, increasing the capacity of our Swin backbone generally improves detection performance (Table~\ref{tab:det_bevformer}). NDS increases steadily from Swin-T to Swin-H, while mAP follows the same overall trend with only a small plateau around Swin-L. Our Swin-L remains close to our Swin-H, and a similar pattern is observed in monocular depth estimation (Sec.~\ref{sec:exp_depth}). At matched Swin scales, our representation is competitive with or better than those learned with ImageNet-22K and SimMIM.

\begin{table*}[t]
\centering
\caption{Quantitative evaluation of BEVFormer V2~\cite{yang2023bevformer} on the nuScenes validation set at an input resolution of $1600\times900$. Rows denote pretraining methods and columns denote Swin backbone variants.}
\label{tab:det_bevformer}
\setlength{\tabcolsep}{1.5pt}
\scriptsize
\resizebox{\textwidth}{!}{%
\begin{tabular}{@{}l*{10}{>{\centering\arraybackslash}p{4.4em}}@{}}
\hline
Pretraining & \multicolumn{2}{c}{Swin-T} & \multicolumn{2}{c}{Swin-S} & \multicolumn{2}{c}{Swin-B} & \multicolumn{2}{c}{Swin-L} & \multicolumn{2}{c}{Swin-H} \\
& NDS $\uparrow$ & mAP $\uparrow$ & NDS $\uparrow$ & mAP $\uparrow$ & NDS $\uparrow$ & mAP $\uparrow$ & NDS $\uparrow$ & mAP $\uparrow$ & NDS $\uparrow$ & mAP $\uparrow$ \\
\hline
ImageNet-22K & 51.69 & 42.12 & 53.62 & 45.22 & 53.98 & 45.48 & 54.59 & 45.91 & -- & -- \\
SimMIM~\cite{xie2022simmim} & -- & -- & -- & -- & 54.03 & 45.18 & 54.98 & 46.52 & -- & -- \\
\hline
Ours & 51.56 & 42.49 & \textbf{54.35} & 45.16 & \textbf{55.69} & \textbf{47.67} & \textbf{55.89} & \textbf{47.59} & 56.92 & 48.87 \\
\hline
\end{tabular}
}%
\end{table*}

We further evaluate the representations at a lower input resolution of $704\times256$ and compare them with ViT-based DINOv2 and DINOv3 models (Table~\ref{tab:det_bevformer_704}). At the S scale, our Swin-S clearly outperforms both DINOv2 and DINOv3. At the B scale, our Swin-B remains competitive with DINOv3 and clearly exceeds DINOv2. At the L scale, our Swin-L still outperforms DINOv2, but DINOv3 achieves the strongest performance. Thus, our method shows the clearest advantage over the DINO baselines with compact backbones, while the gap narrows or reverses as the backbone becomes larger.

\begin{table*}[t]
\centering
\caption{Quantitative evaluation of BEVFormer V2~\cite{yang2023bevformer} on the nuScenes validation set at an input resolution of $704\times256$. Rows denote pretraining methods and columns denote backbone variants. In the shared S/B/L groups, DINOv2 and DINOv3 use ViT backbones, while the other methods use Swin backbones.}
\label{tab:det_bevformer_704}
\setlength{\tabcolsep}{1.5pt}
\scriptsize
\resizebox{\textwidth}{!}{%
\begin{tabular}{@{}l*{8}{>{\centering\arraybackslash}p{5.5em}}@{}}
\hline
Pretraining & \multicolumn{2}{c}{Swin-T} & \multicolumn{2}{c}{ViT-S/Swin-S} & \multicolumn{2}{c}{ViT-B/Swin-B} & \multicolumn{2}{c}{ViT-L/Swin-L} \\
& NDS $\uparrow$ & mAP $\uparrow$ & NDS $\uparrow$ & mAP $\uparrow$ & NDS $\uparrow$ & mAP $\uparrow$ & NDS $\uparrow$ & mAP $\uparrow$ \\
\hline
ImageNet-22K & 47.42 & 36.34 & 48.78 & 38.00 & 50.42 & 40.71 & 50.48 & 40.09 \\
DINOv2~\cite{oquab2023dinov2} & -- & -- & 46.24 & 34.88 & 49.08 & 38.36 & 51.91 & 42.05 \\
DINOv3~\cite{simeoni2025dinov3} & -- & -- & 47.17 & 36.08 & \textbf{52.36} & \textbf{42.68} & \textbf{56.40} & \textbf{48.30} \\
\hline
Ours & 46.52 & 36.45 & \textbf{50.76} & \textbf{40.54} & 51.94 & 42.23 & 52.88 & 43.26 \\
\hline
\end{tabular}
}%
\end{table*}

\subsection{3D Occupancy Prediction}
\label{sec:exp_occupancy}
We evaluate 3D occupancy prediction on the nuScenes validation set using SparseOcc~\cite{liu2024fully}. All experiments use eight input frames and a 24-epoch downstream schedule. Table~\ref{tab:sparseocc} reports RayIoU and $\mathrm{RayIoU}_{1\mathrm{m}}$ across pretraining methods and backbone scales.

As shown in Table~\ref{tab:sparseocc}, occupancy performance generally improves as our backbone scales from Swin-T to Swin-L, which reaches 40.04 RayIoU and 33.99 $\mathrm{RayIoU}_{1\mathrm{m}}$. Scaling further to our Swin-H yields nearly identical performance, showing limited additional gain at the largest scale.

Compared with the DINO baselines, our Swin-S and Swin-B outperform both DINOv2 and DINOv3 on the two reported metrics, demonstrating strong transfer at small and medium scales. At the L scale, our Swin-L remains stronger than DINOv2, while DINOv3 achieves the best performance and leads our model by 0.98 RayIoU and 0.94 $\mathrm{RayIoU}_{1\mathrm{m}}$. Overall, the relative gains are most pronounced at the S and B scales, whereas DINOv3 retains an advantage at the L scale.

\begin{table*}[t]
\centering
\caption{RayIoU and $\mathrm{RayIoU}_{1\mathrm{m}}$ evaluation of SparseOcc~\cite{liu2024fully} on the nuScenes validation set. Rows denote pretraining methods and columns denote backbone variants. In the shared S/B/L groups, DINO uses ViT backbones, while the other methods use Swin backbones.}
\label{tab:sparseocc}
\setlength{\tabcolsep}{1.5pt}
\scriptsize
\resizebox{\textwidth}{!}{%
\begin{tabular}{@{}l*{10}{>{\centering\arraybackslash}p{6.1em}}@{}}
\hline
Pretraining & \multicolumn{2}{c}{Swin-T} & \multicolumn{2}{c}{ViT-S/Swin-S} & \multicolumn{2}{c}{ViT-B/Swin-B} & \multicolumn{2}{c}{ViT-L/Swin-L} & \multicolumn{2}{c}{Swin-H} \\
& RayIoU $\uparrow$ & $\mathrm{RayIoU}_{1\mathrm{m}}\uparrow$ & RayIoU $\uparrow$ & $\mathrm{RayIoU}_{1\mathrm{m}}\uparrow$ & RayIoU $\uparrow$ & $\mathrm{RayIoU}_{1\mathrm{m}}\uparrow$ & RayIoU $\uparrow$ & $\mathrm{RayIoU}_{1\mathrm{m}}\uparrow$ & RayIoU $\uparrow$ & $\mathrm{RayIoU}_{1\mathrm{m}}\uparrow$ \\
\hline
ImageNet-22K & 35.50 & 29.40 & 36.80 & 30.40 & 37.60 & 31.30 & 37.60 & 31.40 & -- & -- \\
SimMIM~\cite{xie2022simmim} & -- & -- & -- & -- & 38.00 & 31.70 & 38.60 & 32.60 & -- & -- \\
DINOv2~\cite{oquab2023dinov2} & -- & -- & 35.90 & 29.50 & 37.10 & 31.00 & 39.00 & 32.80 & -- & -- \\
DINOv3~\cite{simeoni2025dinov3} & -- & -- & 36.73 & 30.40 & 38.79 & 32.65 & \textbf{41.02} & \textbf{34.93} & -- & -- \\
\hline
Ours & 36.72 & 30.50 & \textbf{38.45} & \textbf{32.40} & \textbf{39.41} & \textbf{33.13} & 40.04 & 33.99 & 40.03 & 33.97 \\
\hline
\end{tabular}
}%
\end{table*}

\subsection{End-to-End Planning}
\label{sec:exp_planning}
Beyond perception, we evaluate whether our object-centric representation transfers to end-to-end planning. We integrate our Swin-L encoder into DriveSuprim~\cite{yao2026drivesuprim} and evaluate it under NAVSIMv2~\cite{cao2025pseudosimulation}, which extends the original NAVSIM benchmark~\cite{dauner2024navsim}. Following the setup of Xu~et~al.~\cite{xu2026geometry}, all compared variants use the same planning decoder and training configuration and differ only in the visual backbone. We report the extended predictive driver model score (EPDMS) using the corrected NAVSIM evaluator.\footnote{The evaluator correction is documented in \url{https://github.com/autonomousvision/navsim/issues/151}.}

As shown in Table~\ref{tab:planning_navsimv2}, our Swin-L obtains 88.9 EPDMS, outperforming DINOv2-L by 1.7 points and remaining within 0.1 points of DINOv3-L. DA-ViT-L achieves the highest aggregate score of 90.5, leaving a gap of 1.6 points to our model. This gap may partly reflect the explicit depth supervision used by DA-ViT-L, which provides direct scene-geometry cues that are absent from our object-grouping objective. Incorporating complementary depth supervision while retaining motion-derived object grouping is a promising direction for future work.

%^At the metric level, our model obtains the best no-at-fault collision (NC) score of 98.6 and the best time-to-collision (TTC) score of 98.1 among the compared backbones. It also matches the best history-comfort score of 98.3 and improves extended comfort over DINOv2-L and DINOv3-L. In contrast, DA-ViT-L performs better on drivable-area compliance, ego progress, lane keeping, and extended comfort, which helps explain its higher aggregate EPDMS. Overall, our model transfers competitively to end-to-end planning, with its strongest relative performance concentrated on collision-related metrics.

\begin{table*}[t]
\centering
\caption{Planning results on NAVSIMv2 using the corrected NAVSIM evaluator. Baseline results are from Xu~et~al.~\cite{xu2026geometry}; the Ours row is our evaluation. All variants use the same DriveSuprim planning decoder and training configuration and differ only in the visual backbone. NC: no-at-fault collision; DAC: drivable-area compliance; DDC: driving-direction compliance; TL: traffic-light compliance; EP: ego progress; TTC: time-to-collision; LK: lane keeping; HC: history comfort; EC: extended comfort.}
\label{tab:planning_navsimv2}
\setlength{\tabcolsep}{4.6pt}
\footnotesize
\begin{tabular}{lcccccccccc}
\hline
Backbone & NC$\uparrow$ & DAC$\uparrow$ & DDC$\uparrow$ & TL$\uparrow$ & EP$\uparrow$ & TTC$\uparrow$ & LK$\uparrow$ & HC$\uparrow$ & EC$\uparrow$ & EPDMS$\uparrow$ \\
\hline
DA-ViT-L~\cite{yang2024depth} & 98.4 & \textbf{98.6} & \textbf{99.6} & \textbf{99.8} & \textbf{90.5} & 97.8 & \textbf{97.0} & \textbf{98.3} & \textbf{78.6} & \textbf{90.5} \\
DINOv2-L~\cite{oquab2023dinov2} & 97.9 & 97.0 & 99.3 & 99.7 & 87.7 & 97.0 & 95.5 & 98.2 & 77.1 & 87.2 \\
DINOv3-L~\cite{simeoni2025dinov3} & 98.1 & 97.8 & 99.5 & 99.7 & 89.5 & 97.4 & 96.2 & \textbf{98.3} & 77.3 & 89.0 \\
Ours, Swin-L & \textbf{98.6} & 97.1 & 99.5 & 99.7 & 89.4 & \textbf{98.1} & 96.2 & \textbf{98.3} & 77.6 & 88.9 \\
\hline
\end{tabular}
\end{table*}

\subsection{Effectiveness of the Complete Cycle-2 Pipeline}
\label{sec:exp_mvst_effect}
We evaluate the complete Cycle-2 pipeline by comparing the Cycle-1 Swin-H encoder with our final Cycle-2 Swin-H under the same downstream settings. As summarized in Table~\ref{tab:mvst_downstream}, the complete Cycle-2 pipeline improves all primary metrics: SILog decreases from 5.90 to 5.69 on KITTI depth estimation, NDS increases from 56.80 to 56.92 on nuScenes 3D detection, and RayIoU increases from 39.71 to 40.03 on nuScenes occupancy prediction.

\begin{table*}[t]
\centering
\caption{Complete Cycle-2 pipeline comparison under the Swin-H architecture. The Cycle-2 row reports our final Swin-H and therefore includes the motion-verified pseudo-label set, continued Cycle-2 optimization, and the final same-architecture feature-distillation stage. The pretraining-set sizes count pseudo-labeled frames. Downstream architectures, training schedules, and evaluation protocols are identical between rows.}
\label{tab:mvst_downstream}
\setlength{\tabcolsep}{5pt}
\footnotesize
\begin{tabular}{lllccc}
\hline
Pipeline & Pretraining supervision & Model & KITTI SILog $\downarrow$ & nuScenes NDS $\uparrow$ & nuScenes RayIoU $\uparrow$ \\
\hline
Cycle-1 & Motion-derived (195M) & Cycle-1 Swin-H & 5.90 & 56.80 & 39.71 \\
Cycle-2 & Motion-verified (421M) & Ours (Swin-H) & \textbf{5.69} & \textbf{56.92} & \textbf{40.03} \\
\hline
\end{tabular}
\end{table*}

\textbf{Control for Cycle-2 supervision.}
The Swin-H comparison above establishes the combined effect of the complete Cycle-2 pipeline. To isolate whether the improved Cycle-2 pseudo-label set, rather than simply longer direct pretraining on the original Cycle-1 labels, accounts for a transfer gain, we conduct a smaller controlled experiment with Swin-T. We compare the 50-epoch Cycle-1 model with a longer 75-epoch Cycle-1 run and with a Cycle-2 branch initialized from the 50-epoch Cycle-1 checkpoint and trained for 10 epochs on the motion-verified labels. The backbone, optimizer, representation objective, augmentations, batch size, and input resolution are held fixed across the two continuation branches. All three encoders are transferred using exactly the same DCDepth architecture, training schedule, and evaluation protocol. The additional Cycle-1 continuation processes approximately 4.88B samples ($25\times195$M), whereas the Cycle-2 branch processes approximately 4.21B samples ($10\times421$M), giving Cycle-2 a slightly smaller additional pretraining budget. This direct-pretraining control intentionally excludes the final Swin-H feature-distillation stage.

As shown in Table~\ref{tab:mvst_swin_t_control}, extending Cycle-1 pretraining from 50 to 75 epochs does not yield a consistent improvement. In contrast, 10 epochs of Cycle-2 pretraining improve every non-saturated depth metric. These results isolate the benefit of the improved quality and coverage of the Cycle-2 pseudo-labels relative to further optimization on the original labels; they do not separately quantify the contribution of the final Swin-H feature-distillation stage.

\begin{table*}[t]
\centering
\caption{Swin-T direct-pretraining control for Cycle-2 supervision on KITTI depth estimation with DCDepth. Additional sample exposures are measured relative to the 50-epoch Cycle-1 reference. The Cycle-1 continuation and Cycle-2 branch use comparable additional pretraining budgets, while all downstream settings are identical. This control intentionally excludes the final Swin-H feature-distillation stage.}
\label{tab:mvst_swin_t_control}
\setlength{\tabcolsep}{3.6pt}
\scriptsize
\begin{tabular}{llccccccc}
\hline
Variant & Pretraining schedule & Additional exposures & Abs Rel $\downarrow$ & Sq Rel $\downarrow$ & RMSE $\downarrow$ & RMSE log $\downarrow$ & $\delta_1\uparrow$ & SILog $\downarrow$ \\
\hline
Cycle-1 & 50 epochs on Cycle-1 labels & -- & 0.0509 & 0.1409 & 2.0084 & 0.0769 & 0.976 & 6.9709 \\
Cycle-1, longer & 75 epochs on Cycle-1 labels & 4.88B & 0.0512 & 0.1446 & 2.0052 & 0.0773 & 0.973 & 6.9885 \\
Cycle-2 & 50 Cycle-1 + 10 Cycle-2 epochs & 4.21B & \textbf{0.0498} & \textbf{0.1385} & \textbf{1.9766} & \textbf{0.0756} & \textbf{0.978} & \textbf{6.8761} \\
\hline
\end{tabular}
\end{table*}

\textbf{Pseudo-label coverage and agreement.}
Figure~\ref{fig:cycle_label_coverage} qualitatively illustrates the label expansion produced by Motion-Verified Self-Training. In these examples, conservative flow clustering labels only a subset of the visible instances, whereas Cycle-2 recovers additional object regions by combining model proposals with motion evidence. To examine whether this expanded coverage comes from simply accepting noisier proposals, we compare the Cycle-1 and Cycle-2 pseudo-labels with masks generated independently by SAM~2~\cite{ravi2024sam2}. SAM~2 is used only as a reference for this analysis and does not provide ground truth or training supervision. We do not expect exact instance-level agreement, as our motion-derived pseudo-labels and the automatic masks generated by SAM~2 may partition the same foreground regions at different granularities. We sample 200 frames labeled by both cycles and 200 frames labeled only in Cycle-2, using one frame from each of 400 distinct videos.

As shown in Table~\ref{tab:pseudo_label_agreement}, Cycle-2 increases the average number of pseudo-instances from 1.6 to 2.4 and reference-foreground coverage from 13.8\% to 16.6\% on the paired frames. The fraction of frames containing at least one matched instance also rises from 45.0\% to 54.0\%, while IoU changes from 83.8\% to 81.6\%. On the Cycle-2-only frames, IoU reaches 81.2\%, and 47.5\% of frames contain at least one match. Together with the qualitative examples, these results indicate that Cycle-2 expands pseudo-label coverage while retaining high overlap with the independent reference.

\begin{table*}[t]
\centering
\caption{Pseudo-label coverage and agreement with independently generated SAM~2 reference masks. Ref. coverage measures the reference foreground covered by pseudo-labels.}
\label{tab:pseudo_label_agreement}
\setlength{\tabcolsep}{8pt}
\footnotesize
\begin{tabular}{llcccc}
\hline
Subset & Labels & Inst./frame & Ref. coverage (\%) & IoU (\%) & Matched frames (\%) \\
\hline
Paired frames & Cycle-1 & 1.6 & 13.8 & 83.8 & 45.0 \\
Paired frames & Cycle-2 & 2.4 & 16.6 & 81.6 & 54.0 \\
Cycle-2-only frames & Cycle-2 & 2.1 & 14.6 & 81.2 & 47.5 \\
\hline
\end{tabular}
\end{table*}

\begin{figure*}[t]
\centering
\setlength{\tabcolsep}{1pt}
\begin{tabular}{@{}ccccc@{}}
\footnotesize Input & \footnotesize Optical Flow & \footnotesize Flow Boundaries & \footnotesize Cycle-1 Labels & \footnotesize Cycle-2 Labels \\[2pt]
\includegraphics[width=0.192\textwidth]{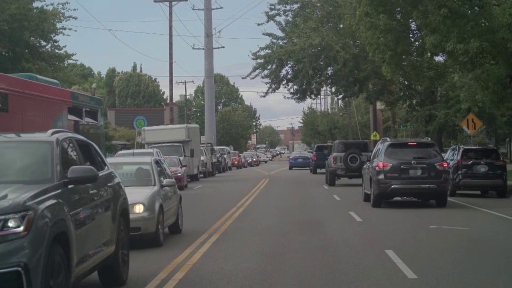} &
\includegraphics[width=0.192\textwidth]{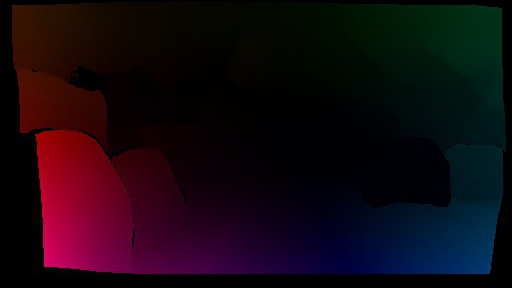} &
\includegraphics[width=0.192\textwidth]{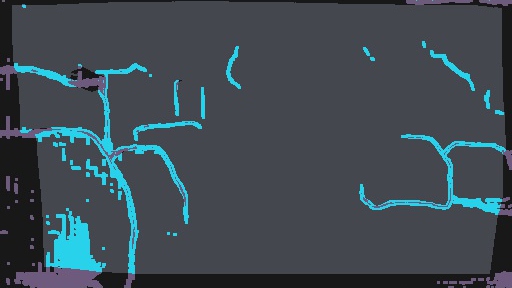} &
\includegraphics[width=0.192\textwidth]{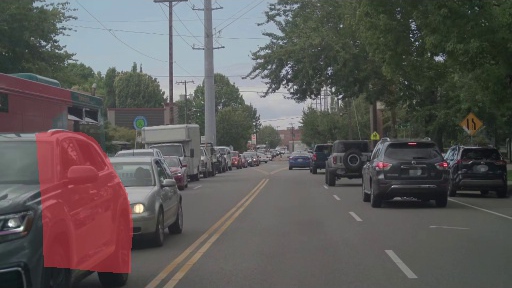} &
\includegraphics[width=0.192\textwidth]{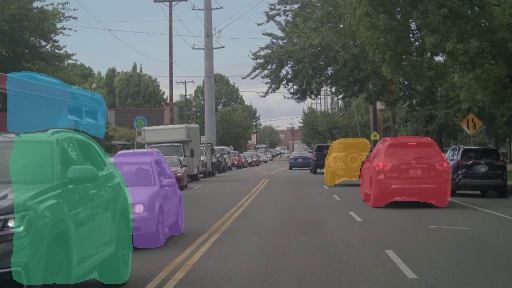} \\[1pt]
\includegraphics[width=0.192\textwidth]{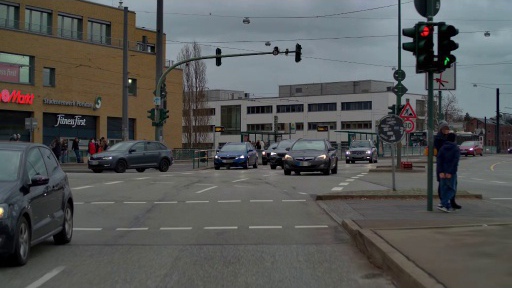} &
\includegraphics[width=0.192\textwidth]{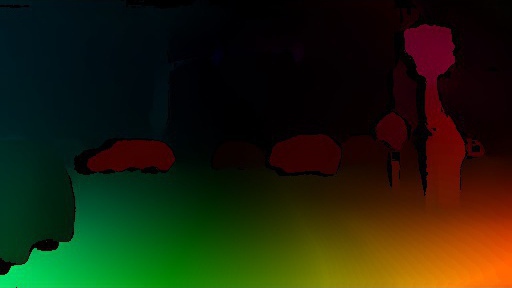} &
\includegraphics[width=0.192\textwidth]{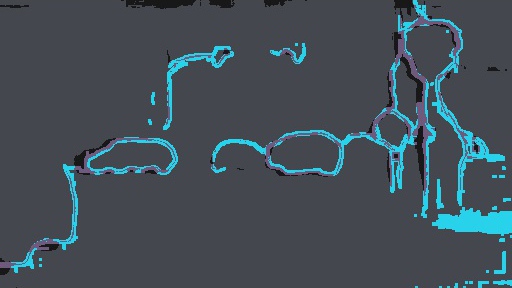} &
\includegraphics[width=0.192\textwidth]{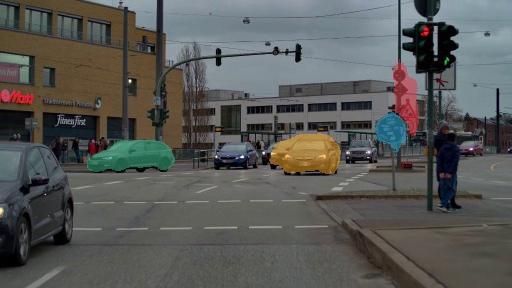} &
\includegraphics[width=0.192\textwidth]{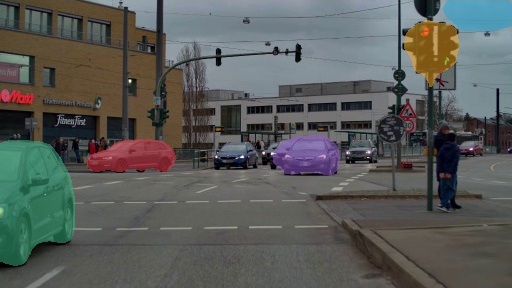} \\[1pt]
\includegraphics[width=0.192\textwidth]{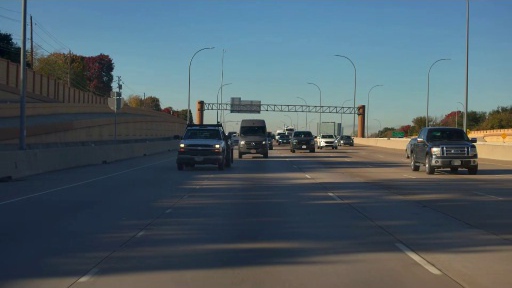} &
\includegraphics[width=0.192\textwidth]{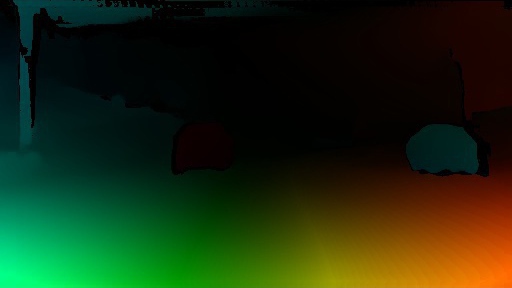} &
\includegraphics[width=0.192\textwidth]{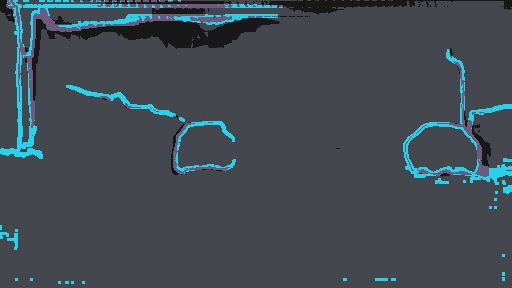} &
\includegraphics[width=0.192\textwidth]{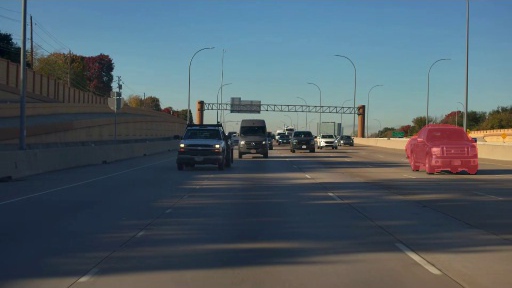} &
\includegraphics[width=0.192\textwidth]{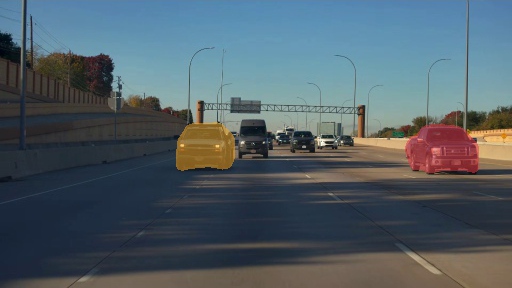}
\end{tabular}
\caption{Qualitative comparison of Cycle-1 and Cycle-2 pseudo-labels. Each row shows an input frame, its optical flow, flow boundaries used as motion evidence, and the pseudo-labels generated in the two cycles. Conservative flow clustering in Cycle-1 captures only instances supported by sufficiently complete motion regions, whereas Cycle-2 combines model proposals with motion-based verification to recover additional instances. Colors distinguish instances within each label panel and do not denote semantic categories.}
\label{fig:cycle_label_coverage}
\end{figure*}

\subsection{Complementarity to Appearance-Based Pretraining}
\label{sec:exp_complementarity}
To examine whether motion-derived and appearance-based representations provide complementary cues, we fuse their backbone features before the DCDepth decoder. Specifically, features from the two encoders are concatenated along the channel dimension. As shown in Table~\ref{tab:depth_dino}, combining our Swin-T representation with DINOv2 ViT-S improves SILog from 6.714 to 6.622. More importantly, under the same two-encoder architecture and fusion design, Ours+DINO outperforms ImageNet+DINO, reducing SILog from 7.071 to 6.622 and RMSE from 2.117 to 1.948. This controlled comparison indicates that our motion-derived representation provides cues complementary to the appearance-based representation learned by DINOv2.

\begin{table}[t]
\centering
\caption{Feature-level complementarity on the KITTI Eigen split using DCDepth~\cite{NEURIPS2024_76bea0a1}. Ours and IN use Swin-T, while DINO uses DINOv2 ViT-S. For dual-encoder variants, the two feature maps are concatenated along the channel dimension before the depth decoder.}
\label{tab:depth_dino}
\setlength{\tabcolsep}{2.1pt}
\scriptsize
\begin{tabular}{ccc|ccc}
\hline
Ours & DINO & IN & SILog $\downarrow$ & Abs Rel $\downarrow$ & RMSE $\downarrow$ \\
\hline
\checkmark & & & \underline{6.714} & \underline{0.049} & \underline{1.958} \\
& \checkmark & & 7.119 & 0.052 & 2.153 \\
& & \checkmark & 7.455 & 0.055 & 2.182 \\
\hline
& \checkmark & \checkmark & 7.071 & 0.052 & 2.117 \\
\checkmark & \checkmark & & \textbf{6.622} & \textbf{0.048} & \textbf{1.948} \\
\hline
\end{tabular}
\end{table}

\section{Discussion}
\label{sec:discussion}

Our framework depends on the quality of optical flow. The forward-backward consistency check and subsequent filtering remove many unreliable estimates, but they cannot recover motion evidence that is absent from the estimated flow field. Fast motion, severe occlusion, motion blur, weak texture, non-rigid deformation, and incomplete motion boundaries can all prevent the Cycle-1 clustering pipeline from producing reliable pseudo-instances. Motion-Verified Self-Training improves the utilization of incomplete motion cues by combining model proposals with flow-based verification and refinement, but it cannot recover an object boundary when neither the flow field nor the Cycle-1 encoder provides sufficient evidence.

This dependence also introduces a selection bias into the pretraining data. The pseudo-labeling pipeline preferentially retains frames with reliable flow and objects whose boundaries are sufficiently evident in the estimated flow. Small objects, heavily occluded regions, and objects whose motion is difficult to distinguish from their surroundings are less likely to receive valid pseudo-labels. Because the Cycle-2 encoder is initialized from Cycle-1 supervision, it may inherit some of these biases even as it increases label coverage. %Moreover, processing hundreds of millions of frames requires substantial computation for optical-flow estimation, mask construction, and filtering.

Another phenomenon worth studying is that the performance gain from scaling from Swin-L to Swin-H seems to diminish. We attribute this to the insufficient amount and diversity of data. We leave further scaling of motion-induced representation learning to future work.
%The learned representation also reflects the granularity of its supervision. Encouraging feature consistency over an entire pseudo-instance benefits object-level grouping, but may suppress distinctions among parts within the same object. Consequently, the representation may be less suitable for tasks that require fine-grained local correspondence or part-level discrimination. Local correspondence objectives could be combined with our object-level objective to preserve both object unity and internal structure. %In addition, although our supervision is derived from videos, the resulting encoder operates on individual frames and does not explicitly maintain object identity over time, model interactions among objects, or predict future dynamics. Extending the representation with temporal association and prediction remains an important direction.

In the current framework, the pretrained optical-flow model is treated as a fixed supervision generator. A promising direction is to build a joint low-level motion and high-level instance learning framework. On the one hand, the learned instance prior could improve optical flow in weakly textured or partially occluded regions and reduce erroneous flow propagation across object boundaries. On the other hand, better flow estimation could benefit instance learning. This complementary learning framework offers a scalable path toward increasingly reliable supervision and richer object-centric representations learned from raw video. %establish a motion-to-object-to-motion bootstrapping loop. Optical flow would first generate pseudo-instance labels for learning an object-centric encoder; the resulting encoder could then initialize or augment the flow model with object-level features and boundary priors. Such priors may improve correspondence in weakly textured or partially occluded regions and reduce erroneous flow propagation across object boundaries. The improved flow model could subsequently generate more reliable motion supervision for the next generation of object-centric encoders. This iterative process could allow motion estimation and object representation learning to improve each other, offering a scalable path toward increasingly reliable supervision and richer object-centric representations learned from raw video.

\section{Conclusion}
\label{sec:conclusion}
We presented a biologically inspired framework for learning object-centric visual representations from raw videos without human annotations on the target video corpus or camera calibration. The framework converts optical flow into category-agnostic pseudo-instance supervision and transfers the resulting object-level organization to a single-image encoder through dense pairwise metric learning. By scaling the unified processing pipeline to 7,163 hours of heterogeneous video, we obtained 195M motion-derived pseudo-labeled frames. Motion-Verified Self-Training then combined appearance-based proposals with proposal-independent motion evidence to recover additional supervision from the same videos, expanding the training set to 421M frames. We further scaled the representation to a Swin-H backbone and distilled it into a family of Swin backbones.

Across monocular depth estimation, 3D object detection, 3D occupancy prediction, and end-to-end planning, the learned representations transfer competitively with supervised and self-supervised alternatives, with particularly strong transfer on geometry- and instance-sensitive tasks. %The controlled Swin-T comparison further demonstrates that Cycle-2 supervision yields better depth transfer than simply extending pretraining on the Cycle-1 labels.
These results support the view that motion provides a distinct form of visual supervision: it encourages object unity and instance separation without specifying semantic categories. Motion-derived pretraining therefore complements, rather than replaces, category-oriented and temporal objectives. Combining these sources of supervision, while improving the reliability and coverage of motion cues, offers a promising direction toward more complete visual representations learned from raw video.

\bibliographystyle{IEEEtran}
\bibliography{refs}

\begin{IEEEbiography}[{\includegraphics[width=1in,height=1.25in,clip,keepaspectratio]{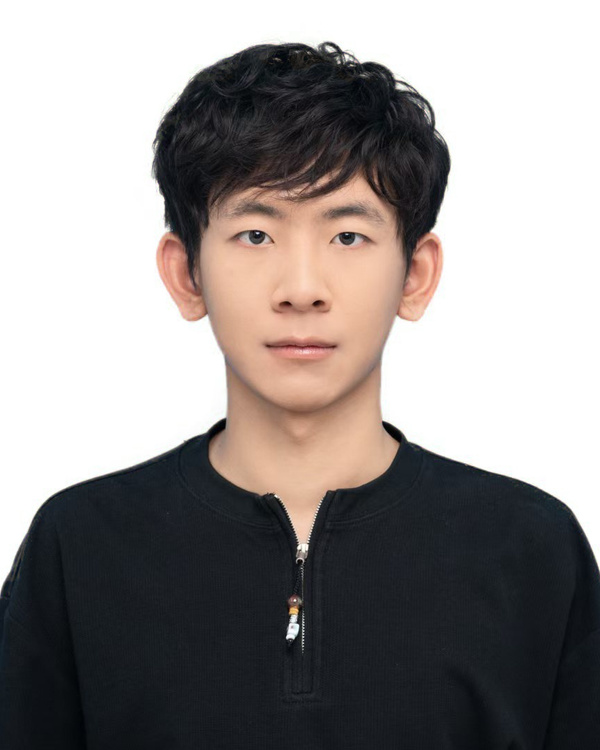}}]{Boshi Li}
received the B.E. degree in electronic information engineering from Beijing University of Posts and Telecommunications, Beijing, China, in 2026. He is currently pursuing the master's degree with the School of Computer Science, Beijing University of Posts and Telecommunications. His research interests include computer vision, self-supervised representation learning, object-centric representation learning, and autonomous driving.
\end{IEEEbiography}

\begin{IEEEbiography}[{\includegraphics[width=1in,height=1.25in,clip,keepaspectratio]{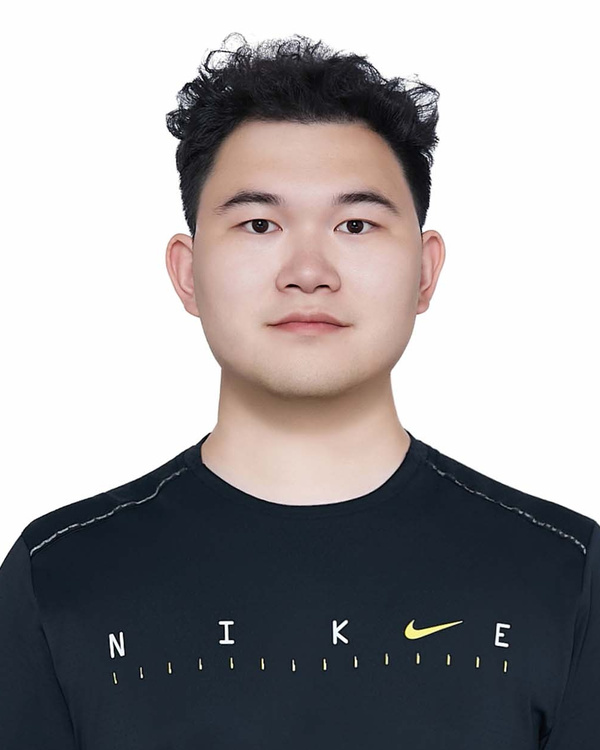}}]{Xiaohui Wang}
received the B.E. and M.E. degrees in computer science and technology from Beijing University of Posts and Telecommunications, Beijing, China, in 2023 and 2026, respectively. His research interests include computer vision, visual representation learning, self-supervised learning, 3D perception, and autonomous driving.
\end{IEEEbiography}

\begin{IEEEbiography}[{\includegraphics[width=1in,height=1.25in,clip,keepaspectratio]{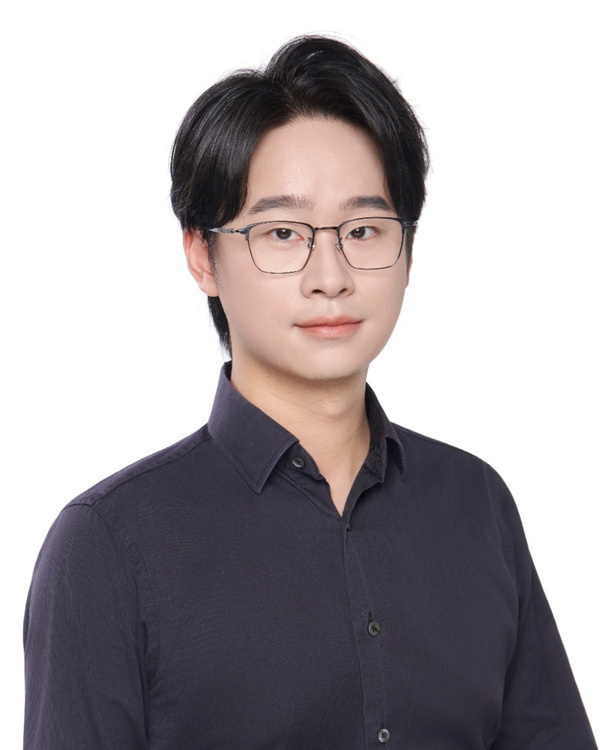}}]{Xiaoyang Wu}
received the Ph.D. degree in computer science from The University of Hong Kong, where he was advised by Prof. Hengshuang Zhao. He is currently a Postdoctoral Associate at New York University, working with Prof. Saining Xie. His research focuses on 3D and 4D perception, spatial representation learning, and embodied intelligence, with an emphasis on learning generalizable representations of the physical world.
\end{IEEEbiography}

\begin{IEEEbiography}[{\includegraphics[width=1in,height=1.25in,clip,keepaspectratio]{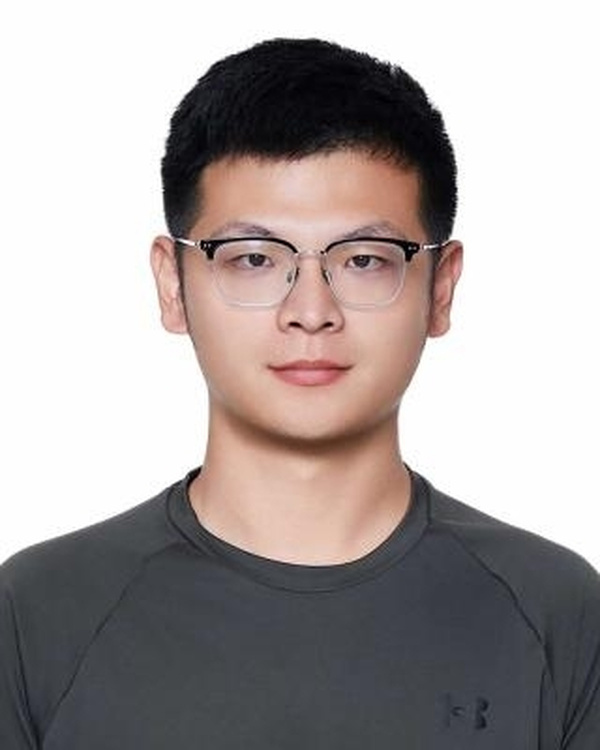}}]{Zhichao Li}
received the B.E. and M.E. degrees in communication engineering from Beijing University of Posts and Telecommunications, Beijing, China, in 2015 and 2018, respectively. From 2017 to 2024, he was a Researcher with TuSimple, where he worked on perception and learning algorithms for autonomous driving. Since 2024, he has been a Senior Researcher with Xiaomi EV, Beijing, China. His research interests include embodied AI, large language models, representation learning, and world models.
\end{IEEEbiography}

\begin{IEEEbiography}[{\includegraphics[width=1in,height=1.25in,clip,keepaspectratio]{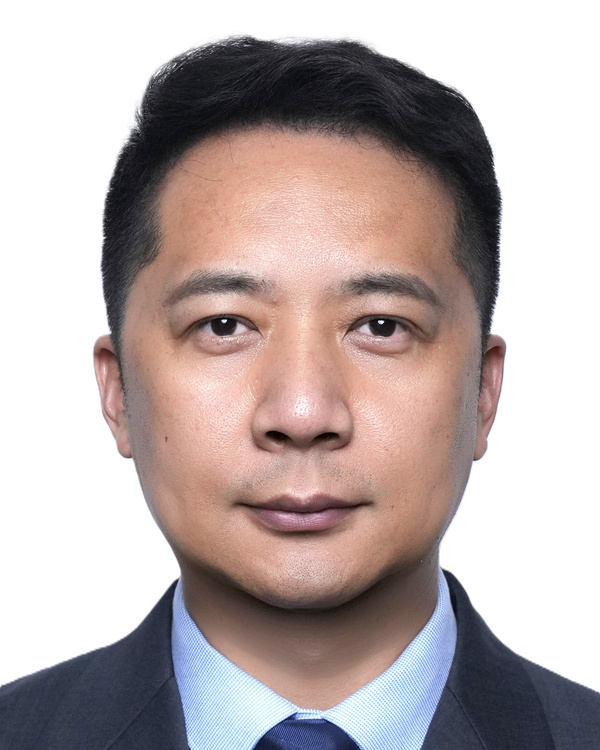}}]{Ya Yang}
received the master's degree from Beijing University of Posts and Telecommunications (BUPT), Beijing, China, in 1999. He is currently an Associate Professor with the School of Computer Science, BUPT, where he is also the Head Coach of the BUPT ICPC training team. His research interests include autonomous driving and the application of large language models to programming competitions.
\end{IEEEbiography}

\begin{IEEEbiography}[{\includegraphics[width=1in,height=1.25in,clip,keepaspectratio]{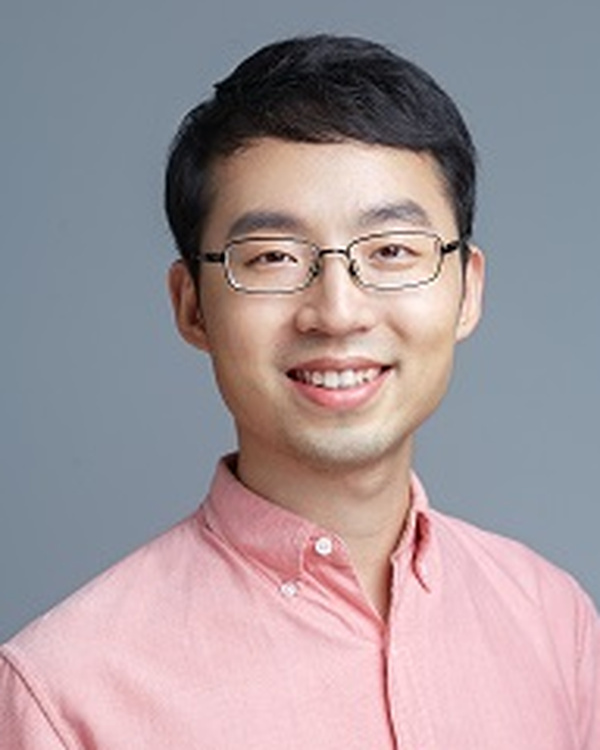}}]{Naiyan Wang}
received the B.S. degree from Zhejiang University, Hangzhou, China, in 2011, and the Ph.D. degree from the Department of Computer Science and Engineering, The Hong Kong University of Science and Technology, Hong Kong, in 2015. He is currently a Distinguished Scientist with Xiaomi EV. Previously, he was the Chief Scientist of TuSimple and the CTO of TuSimple China, where he led the development of Level-4 autonomous trucks. His research interests include computer vision and machine learning and their applications in autonomous driving and robotics.
\end{IEEEbiography}

\end{document}